\documentclass{article}

\usepackage[preprint]{tmlr}

\usepackage{hyperref}
\usepackage{url}
\usepackage{graphicx}
\usepackage{booktabs}
\usepackage{multirow}
\usepackage{amsmath,amssymb}
\usepackage{enumitem}
\usepackage{microtype}

\graphicspath{{figures/}}

\title{A Circuit, Not \emph{The} Circuit: Non-Unique Causal
  Localisation of the Mamba-2 State Sink}

\author{%
  \name Yuhang Jiang \email jyhtjtj@gmail.com \\
  \addr Department of Information Engineering and Computer Science \\
  University of Trento
  \AND
  \name Bowen Zhang \email bzhang2111@gmail.com \\
  \addr Department of Information Engineering and Computer Science \\
  University of Trento
}

\begin{document}
\maketitle

\begin{abstract}
Mechanistic interpretability routinely reads a probe and labels its
top-activating units as \emph{the} circuit executing the computation.
We test the move in Mamba, on the \emph{state sink}: the selective
state-space analogue of the Transformer attention sink, where the
$\Delta$-gate fires disproportionately on boundary tokens such as BOS
and newline. At Mamba-1 channel
granularity the probe's units carry the causal effect. At Mamba-2 head
granularity the label-to-locus link breaks in three ways. The causal
set is not unique: a near-disjoint set of top-activation heads, sharing
almost none of the specialists, matches or exceeds their ablation
effect. Representation does not track function: \emph{dual} heads, a
quarter to over a third of all heads, respond alike to BOS and newline
yet do not reproduce the specialists' causal profile. And the effect dissociates
by intervention surface: perturbing the input-dependent $\Delta$ the
probe reads moves the loss by at most $0.16$ nats, while suppressing
the same heads' output moves it by $1.5$ to $7.3$ nats. The heads still
matter behaviourally, dropping needle-retrieval success on a
RULER-style probe from $30/30$ to $0/30$ at $1024$ context on both
flagship checkpoints. A representational signature in
a selective state-space model thus marks a behaviourally important head
set without pinning down the circuit: a single probe recovers a
circuit, not \emph{the} circuit.

\end{abstract}

\section{Introduction}
\label{sec:intro}

When mechanistic interpretability locates a circuit in a language
model, it typically reads a probe and labels the highest-activating
units as \emph{the} circuit. The label is doing real work: it says that
the units carrying a representational signature are the units doing the
computation. We test that move at intervention scale in Mamba, and at
Mamba-2 head granularity it comes apart. The signature and the circuit
are not the same thing.

The setting makes the test sharp. At Mamba-1 channel granularity the
probe's units do carry the causal effect. At Mamba-2 head granularity
the link fails on three counts: a near-disjoint set of heads carries an
effect at least as large; representationally similar heads do not; and
perturbing the $\Delta$ selectivity the probe reads leaves the loss
nearly unchanged, while suppressing the same heads' output moves it by
nats. A representational signature there marks a behaviourally
important head set without pinning down the circuit, so single-probe
localisations warrant re-audit.

The question is not local to Mamba. State-space and hybrid models are
now released at scale, including Jamba~\citep{lieber2024jamba},
Falcon-Mamba~\citep{zuo2024falconmamba},
RecurrentGemma~\citep{botev2024recurrentgemma}, and
Zamba~\citep{glorioso2024zamba}, and the interpretability toolkit built
for Transformers is being carried over to them, with
\citet{paulo2024features} asking directly whether it transfers. Our
result is a caution for that effort: on a head-shared substrate the
signature these methods read does not single out the causal circuit, so
a localisation taken from one probe should be checked causally before
it is relied on. The precedent is the attention sink itself, which
began as an interpretability observation and later mattered for
efficient inference~\citep{xiao2024streamingllm}.

Mamba is a selective state-space
model~\citep{gu2024mamba,dao2024mamba2}. Where a Transformer attends
over all previous tokens, Mamba carries a compressed recurrent state
forward through the sequence. A per-token, input-dependent gate, the
timestep $\Delta$, sets how strongly each token writes into that state:
a large $\Delta$ overwrites the running state, a small one lets it
persist. Mamba-2 differs from Mamba-1 in
grouping the hidden channels into \emph{heads} that share one $\Delta$
gate, so the natural unit of analysis moves from the channel in Mamba-1
to the head in Mamba-2, and we study both.

Decoder-only Transformer language models allocate disproportionate
attention mass to the first token, the
\emph{attention sink}~\citep{xiao2024streamingllm,gu2024attentionsinks,cancedda2024spectral}.
Recent work has made attention sinks both localisable and
architecturally suppressible: \citet{qiu2025gated} report that a
head-specific sigmoid gate after SDPA reduces average first-token
attention from $46.7\%$ to $4.8\%$. We use \emph{state sink} to refer
to the analogous phenomenon in selective state-space language
models~\citep{gu2024mamba,dao2024mamba2}: the disproportionate
allocation of token-conditional gate $\Delta$ activation to boundary
tokens such as BOS and newline that anchors the hidden state. Attention sink is
measured in attention-score space; the state sink is measured in
$\Delta$-gate space. Mechanistically the two are distinct: the
attention sink emerges from the Softmax sum-to-one constraint that
forces residual mass onto a low-information
anchor~\citep{xiao2024streamingllm}, whereas the state sink arises
from the SSM's selective gating, which independently elevates $\Delta$
at boundary tokens with no normalisation trade-off. Related SSM work
supplies adjacent
operationalisations~\citep{chiang2025quamba,ye2025longmamba},
including activation-outlier magnitude and hidden-state survival.

The state sink is straightforward to detect; the difficulty is
localising it. A single-bucket probe takes the top-activating heads on
the BOS bucket of the $\Delta$-gate and recovers a small set of
\emph{bos-specialist} heads, about $5\%$ at 2.7B, whose ablation has
large causal effects on BOS-context predictions. That set is not the
unique causal locus: a near-disjoint top-activation set, sharing only
$13$ of $254$ heads with it at a Jaccard of $0.03$, produces an effect
at least as large at both scales. Representationally similar heads are
not causally interchangeable either. \emph{Dual} heads, $27$--$35\%$ of
the total, come from the same probe under a multi-class aggregation
rule and are near-collinear with the specialists on BOS and newline,
yet under class-conditional ablation they do not stand in for them. At
Mamba-1 channel granularity the picture is simpler: single-bucket
labels track causal mass more cleanly.

Our evidence is a pre-registered intervention grid ($12{,}096$ cells,
$6{,}048$ evaluated) over four Mamba checkpoints on
wikitext-2~\citep{merity2017wikitext}, with the probe-defined head and
channel sets transferred to a Pile-10k~\citep{gao2020pile} subset for
cross-corpus replication, detailed in \S\ref{sec:causal:method}.
Four controls sharpen the localisation question: a size- and
depth-matched cosine null, alternative probe definitions including a
top-activation set, a threshold sweep, and a joint / marginal
decomposition of the bos-specialist and dual sets. Headline cells are
ported to four Pythia attention
checkpoints~\citep{biderman2023pythia} as a cross-architecture
baseline and to RULER
Needle-in-a-Haystack~\citep{hsieh2024ruler} for retrieval transfer.
The work extends the activation-patching and ablation
tradition~\citep{olsson2022context,conmy2023automated,sharma2024locating}
to selective state-space substrates.

\begin{figure}[t]
  \centering
  \includegraphics[width=\textwidth]{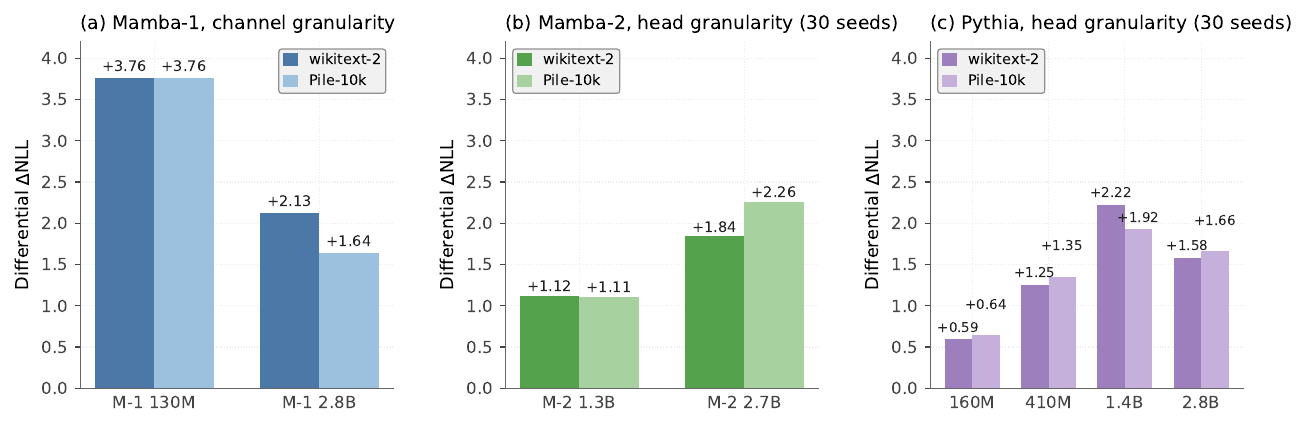}
  \caption{Size-matched bos-specialist causal effect across
    granularities and architectures, under \texttt{gate\_zero}
    ablation. Specialist--complement $\Delta$NLL on the
    first-eight-tokens-after-BOS for (a) Mamba-1 channels,
    (b) Mamba-2 heads (30-seed random-complement bank), and
    (c) Pythia attention heads. Bars: wikitext-2 (dark),
    Pile-10k (paler). Labels in nats.}
  \label{fig:selectivity}
\end{figure}

\paragraph{Findings.} Five findings, each cross-dataset, cross-scale,
or cross-architecture replicated; full numbers in \S\ref{sec:causal}.

\begin{itemize}[itemsep=1pt,topsep=1pt,leftmargin=1.2em]
  \item \textbf{The state sink is a robust phenomenon.} Boundary tokens
    receive disproportionate $\Delta$-gate activation across all four
    checkpoints, measured on wikitext-2; the specialist effect stays
    positive across the threshold sweep at 2.7B and transfers to a
    Pile-10k subset.
  \item \textbf{Causal localisation is non-unique.} A near-disjoint
    top-activation set that shares almost none of the specialists
    matches or exceeds the bos-specialist BOS-context effect at both
    scales, so no single probe's set is the unique locus. The
    bos-specialist and dual sets, where they differ, are not redundant:
    their joint ablation is super-additive on the 2.7B newline target
    (\S\ref{sec:causal:m2_func_decomp}).
  \item \textbf{Representational similarity does not certify function.}
    Dual heads' BOS--newline collinearity of $\cos = 0.89$--$0.90$ sits
    only modestly above a matched null at $0.82$--$0.86$, and the dual
    set does not reproduce the bos-specialist causal profile under
    class-conditional ablation.
  \item \textbf{The effect dissociates by intervention surface.}
    At Mamba-2 heads, perturbing the input-dependent $\Delta$ the probe
    reads moves the loss by at most $0.16$ nats, while suppressing the
    same heads' output moves it by $1.5$ to $7.3$; the intervals are
    disjoint by more than a nat (\S\ref{sec:causal:m2_mechanism}).
  \item \textbf{Downstream reach and architectural correlate.}
    bos-specialist ablation drops needle-retrieval success on a
    RULER-style probe from $30/30$ to $0/30$ at $1024$ context on
    Mamba-1 2.8B and Mamba-2 2.7B (\S\ref{sec:causal:niah}). A random
    channel-bucketing control is consistent with Mamba-2's head-shared
    $\Delta$ projection (\S\ref{sec:causal:t3}); four Pythia checkpoints
    show single-bucket selectivity without comparable multi-set
    structure (\S\ref{sec:causal:pythia}).
\end{itemize}

\paragraph{Why Mamba-2 heads.} Mamba-1's channel-independent $\Delta$
projections let specialist sub-populations on different boundary
classes co-exist, so a single-bucket probe tracks the causal set.
Sharing the $\Delta$ projection across a head's channels could push
each Mamba-2 head to multiplex across boundary
classes~\citep{gu2024mamba,dao2024mamba2}; the boundary-token causal
support then spreads across overlapping head sets that no single-bucket
probe recovers in full.

\section{Related Work}
\label{sec:related}

\paragraph{Attention sinks and BOS-token behaviour.}
Autoregressive Transformers can allocate disproportionate
attention mass to initial tokens, the attention-sink
phenomenon~\citep{xiao2024streamingllm}; subsequent work
characterises sink emergence during
training~\citep{gu2024attentionsinks}, links it to spectral
filters and ``dark signals''~\citep{cancedda2024spectral},
proposes causal-mask
modifications~\citep{yin2024stablemask}, and connects
sink-like behaviour to massive activations, head-level
no-op/outlier patterns, and register tokens in vision
Transformers~\citep{sun2024massive,bondarenko2023quantizable,darcet2024registers}.
\citet{qiu2025gated} show that a head-specific sigmoid gate after
SDPA mitigates attention sinks, reducing average first-token
attention from $46.7\%$ to $4.8\%$.
Work in this thread measures sink behaviour in attention-score
space, typically by attention mass assigned to BOS or other
initial positions; we test the analogous label-to-locus move in
selective state-space models, where the token-conditional gate
$\Delta$ has not been audited for sink-locus identification.
Prior probes of sink-like behaviour operate at the activation
distribution or hidden-state level~\citep{chiang2025quamba,ye2025longmamba};
we add unit-level resolution through channel- and head-targeted
interventions.

\paragraph{Sink mass in selective state-space models.}
\citet{chiang2025quamba} identify SSM-input activation outliers
(high-percentile of $|x|$) for post-training quantisation;
\citet{ye2025longmamba} study hidden-state memory decay and
receptive-field limits in Mamba, using training-free token
filtering to enlarge effective receptive fields. We complement
these with two analytic baselines
computed directly from Mamba's state: a right-tail
$\Pr(\Delta_{\text{pre}} > 0.7)$ statistic for ``reset''
propensity, and a per-channel $-\log A$ baseline derived from the
stored $A_{\log}$ parameter, the structural quantity
governing state-decay in the SSM--attention duality
\citep{dao2024mamba2}. Whether thresholded sink sets agree across
methods, and whether they predict causal effects, have not been
audited at scale; we report a four-method cross-validation in
Appendix~\ref{sec:appendix:phenomenology_extra}.

\paragraph{Mamba mechanistic interpretability and probes.}
\citet{ali2024hidden} reformulate Mamba's selective scan as a
data-dependent attention matrix; \citet{paulo2024features} test
whether selected Transformer interpretability methods transfer to
RNNs/Mamba; \citet{sharma2024locating} adapt the
ROME~\citep{meng2022rome} pipeline to Mamba and localise factual
associations at mid-network SSM layers; and
\citet{ensign2024investigating} adapt IOI circuit analysis and
positional edge attribution patching to Mamba.
\citet{rezaei2024mambalrp} bring layer-wise relevance
propagation to Mamba for input attribution;
\citet{pitorro2025latim} measure latent token-to-token interactions
across Mamba-1 and Mamba-2 layers; and \citet{mohan2026interpreting}
identify activation-subspace bottlenecks and steer SSM behaviour
through them. Linear probes are
standard for representational analysis
\citep{tenney2019bert,belinkov2022probing,hewitt2019control,voita2020information,
ravichander2021probing,elazar2021amnesic};
causal-intervention work cautions that readable representations
need not be behaviourally used and provides methods for testing
that gap
\citep{geiger2021causal,vig2020mediation,heimersheim2024patching,belrose2023leace}.
Our intervention grid perturbs the token-conditional $\Delta$
gate that defines selective state-space
models~\citep{gu2024mamba}; the broader
ablation, activation-patching, and automated-circuit-discovery
tradition is established in
\citep{wang2023interpretability,olsson2022context,vig2020mediation,
meng2022rome,heimersheim2024patching,conmy2023automated,
cunningham2024sparse,templeton2024scaling,marks2024sparse,wu2025retrieval}.
Where prior probe-causality work showed that \emph{what} a
representation encodes is hard to read, we show that
\emph{where} a mechanism lives is equally hard to pin down at
Mamba-2 head granularity, where near-disjoint probes recover the
same causal effect.

\paragraph{Long-context retrieval and SSM limits.}
Pure SSMs have documented weaknesses relative to Transformers on
copying, in-context-learning, and long-context reasoning
tasks~\citep{jelassi2024repeat,waleffe2024empirical};
we use single-key
RULER~\citep{kamradt2023niah,hsieh2024ruler} at $1024$ context
as a behavioural probe of the same causal axis, not as a
benchmark contribution.

\section{Phenomenology}
\label{sec:phenomenology}

\subsection{Specialist classification protocol}
\label{sec:phenomenology:protocol}

For each (layer, channel) in Mamba-1 and (layer, head) in Mamba-2,
we compute mean pre-softplus activation $\bar{a}_{\ell,u,c}$ on
tokens of class
$c \in \{$BOS, newline, punctuation, whitespace\_word$\}$ over a
$30$-document wikitext-2 sample, subtract the unit's mean
activation on a reference (non-class) token set, and threshold the
resulting differential $\Delta\bar{a}_{\ell,u,c}$ at $\tau = 0.5$
(pre-softplus units) to obtain a binary specialty matrix
$S_{\ell, u, c} \in \{0, 1\}$. A unit is a \emph{$c$-specialist}
if exactly one class clears threshold; a \emph{dual} specialist
if two classes clear threshold; a \emph{generalist} if three or
more; \emph{silent} otherwise. The threshold
$\tau = 0.5$ is fixed across all 7 models and pre-registered to
mark the gate's ``on'' state ($\mathrm{softplus}(0.5) \approx
0.97$); values below threshold map to the low-state mode of the
bimodal pre-$\Delta$ distribution
(\S\ref{sec:phenomenology:sinkmass}). Threshold-sensitivity
at $\tau \in \{0.3, 0.5, 0.7, 1.0\}$ is in
Appendix~\ref{sec:appendix:phenomenology_extra}.

\subsection{Per-token-class distributions across scale and architecture}
\label{sec:phenomenology:distributions}

The specialty fraction (bos-specialists / total units, averaged
across layers) scales differently in the two architectures
(Table~\ref{tab:bos_fraction}). In Mamba-1, the fraction grows
log-linearly with parameter count from $19.4\%$ at $130$M to
$29.4\%$ at $2.8$B, with a least-squares fit of
$f = 7.4 \log_{10} N - 41.9$ at $R^{2} = 0.91$. In Mamba-2 the
fraction caps at $4.9$--$8.2\%$ over the same range and moves
non-monotonically, peaking at 1.3B. The asymmetry tracks the absolute
substrate budget. Mamba-1's hidden width grows from
$D_{\text{inner}} = 1{,}536$ to $5{,}120$, leaving room for an
expanding specialist sub-population. Mamba-2's head budget stays
bounded, at $24$ to $80$ heads across the checkpoints we test, and the
per-layer specialist count saturates well below the
channel-granularity rate.

Within a model, the fraction is highly heterogeneous across layers.
At Mamba-1 130M, layer $0$ alone is $76.0\%$ bos-specialist while
the median layer is $0.5\%$; at Mamba-1 2.8B, the most concentrated
layer reaches $92.9\%$ and layer $0$ has fallen to $10.0\%$. The
reorganisation is consistent with BOS handling localising at the
bottom of the network in the smallest model and redistributing
across deeper layers as scale grows
(\S\ref{sec:causal:m1_channel}).

Pythia attention-head bos-specialists, defined in our
operationalisation by mean attention-to-BOS strength~$> 0.10$ and
motivated by the attention-sink practice of measuring first-token
attention mass~\citep{xiao2024streamingllm,gu2024attentionsinks},
account for
$15\%$ to $97\%$ of heads per layer in mid-to-late layers, in line
with prior reports of pervasive first-token sinks in
Transformer LMs~\citep{xiao2024streamingllm}.

\begin{table}[t]
  \centering
  \small
  \begin{tabular}{llrrr}
    \toprule
    Family & Model & units / layer & bos-spec & dual \\
    \midrule
    \multirow{4}{*}{Mamba-1}
      & 130M & $1{,}536$ ch  & $19.4\%$ & $13.8\%$ \\
      & 370M & $2{,}048$ ch  & $20.2\%$ & $21.8\%$ \\
      & 1.4B & $4{,}096$ ch  & $25.0\%$ & $23.8\%$ \\
      & 2.8B & $5{,}120$ ch  & $29.4\%$ & $29.6\%$ \\
    \midrule
    \multirow{3}{*}{Mamba-2}
      & 130M & $24$  heads   & $4.9\%$  & $13.4\%$ \\
      & 1.3B & $64$  heads   & $8.2\%$  & $26.9\%$ \\
      & 2.7B & $80$  heads   & $5.0\%$  & $35.4\%$ \\
    \bottomrule
  \end{tabular}
  \caption{Bos-specialist and dual specialty fractions (mean
    across layers) per model at $\tau = 0.5$ (pre-softplus
    units). The dual fraction converges to $24$--$35\%$ on
    larger-substrate models in both architectures.}
  \label{tab:bos_fraction}
\end{table}

\subsection{Sink-mass cross-validation and \texorpdfstring{$\Delta$}{Delta}-bimodality}
\label{sec:phenomenology:sinkmass}

We compute four per-unit sink-mass
metrics~\citep{chiang2025quamba,ye2025longmamba,gu2024mamba} and
test whether LongMamba state-survival and right-tail
$\Delta_{\text{pre}}$ reset-propensity are negatively correlated
($\rho_{L\Delta} < -0.3$, pre-registered). The prediction holds at \emph{head} granularity in Mamba-2,
with $\rho_{L\Delta} = -0.443$ at 1.3B, $-0.437$ at 2.7B, and a
weaker $-0.282$ at 130M, and fails at \emph{channel} granularity
in Mamba-1, where $\rho_{L\Delta}$ stays in $[-0.02, +0.20]$
across the four scales we test.
The head--vs--channel asymmetry converges with the dual-set
structure at Mamba-2 heads (\S\ref{sec:causal:m2_head}).

A second pre-registered prediction fails, and it bears on what the
state sink is. We predicted that at least $3$ of the $4$ sink-mass
metrics would agree with the probe-derived bos-specialist sets at a
top-$10\%$ Jaccard above $0.4$. Observed agreement is
$0.07$--$0.15$ across all seven models and all four metrics: above the
$\sim\!0.05$ chance level, and far below the prediction
(Appendix~\ref{sec:appendix:phenomenology_extra}). Which units count as
the state sink is therefore method-dependent at the level of the
measurement itself, before any causal question is asked. We report the
phenomenon on the $\Delta$-gate probe that defines it here and treat
the four metrics as related but non-interchangeable operationalisations.
This measurement-level non-uniqueness runs parallel to the causal
non-uniqueness of \S\ref{sec:causal:m2_func_decomp}, where two
near-disjoint probes each turn out to be load-bearing.

A third pre-registered prediction is that the selective-gate
$\Delta$ distribution at sink-relevant positions exhibits
bimodality. A 3-of-4 conjunction
(Hartigan dip~$\geq 0.05$~\citep{hartigan1985dip},
Silverman peak-count $\geq 2$~\citep{silverman1981kernel},
GMM~$\Delta$BIC~$\geq 10$, KDE local-mode count $\geq 2$) holds
on $87.1$--$100.0\%$ of
(layer, bucket) cells across all four tested Mamba checkpoints,
well above the pre-registered $50\%$ floor. Per-model values,
Quamba~$\times$~A-log channel-level breakdown, top-10\% Jaccard
agreement, and the 4-of-4 strict-conjunction calibration
(M-2 2.7B drops to $41.8\%$, with the parametric $\Delta$BIC arm at $47.3\%$)
are in Appendix~\ref{sec:appendix:phenomenology_extra} and
\ref{sec:appendix:bimodality_calibration}.

\section{Causal Analysis}
\label{sec:causal}

\subsection{Method}
\label{sec:causal:method}

The intervention grid is a six-factor sweep: 4 models (Mamba-1
130M / 2.8B, Mamba-2 1.3B / 2.7B) $\times$ 6 interventions
(\texttt{gate\_zero / gate\_one / gate\_mean / delta\_zero /
delta\_median / u\_zero}) $\times$ 7 channel/head scopes $\times$
6 layer scopes $\times$ 4 tasks $\times$ 3 seeds, totalling
$12{,}096$ cells of which $6{,}048$ are evaluated (parity and
Dyck-2 task loaders deferred per pre-registration). Each cell
uses a $30$-document bootstrap CI ($n_{\text{boot}} = 5{,}000$).
We report \emph{differentials}: specialist mean minus
size-matched random-complement mean, since ratios are unstable
when the complement is near zero. The Mamba-2 head-granularity
headline cells additionally use a $30$-seed random-complement bank
under stratified non-adjacent sampling. M-2 head differentials sit
closer to the complement spread than the Mamba-1 channel effect does
(\S\ref{sec:causal:m1_channel}), so they need tighter CIs; the Mamba-1
effect is already stable at the main-grid $3$ seeds. Headline claims rest on the pre-registered cells of
Tables~\ref{tab:m2_30seed} and~\ref{tab:f1_decomp}, so we report
individual $95\%$ bootstrap CIs without grid-wide
multiple-comparison correction. Cells sharing a model, a corpus, or a
probe-selected set are not independent, so counts below report the
conditions in which a result holds, not a number of independent
replications. The grid runs on wikitext-2; the
specialty matrix, and with it every head and channel set, is defined
there. Cross-corpus replication transfers those wikitext-defined sets
unchanged to a filtered Pile-10k subset~\citep{gao2020pile} and
re-scores the headline cells, so Pile-10k tests whether the sets
transfer, not whether the same probe would be recovered from Pile-10k.
The seed protocol, sentinel reproducibility check, and the transferred
cells are in Appendix~\ref{sec:appendix:cross_dataset}.

Intervention labels name hook surfaces, not post-nonlinearity
identities. In Mamba-1, \texttt{gate\_one} sets pre-SiLU
$z=1$ (effective gate $\mathrm{SiLU}(1)$); in Mamba-2 it sets
the gate multiplier to one. In Mamba-2, \texttt{delta\_zero}
zeros the pre-softplus timestep input, leaving the post-softplus
timestep set by the learned bias. We score next-token NLL on two
target classes, BOS-context (the first eight tokens after BOS) and
newline (positions whose target token is a newline), each on
wikitext-2 and Pile-10k.

\paragraph{Roadmap.} \S\ref{sec:causal:m1_channel} establishes
the channel-granularity baseline, where probe specialty tracks causal
mass; \S\ref{sec:causal:m2_head} shows the correspondence breaks at
Mamba-2 head granularity, and \S\ref{sec:causal:m2_func_decomp} shows
the localisation there is non-unique: a near-disjoint probe matches or
exceeds the effect, and representational similarity does not track
function; \S\ref{sec:causal:m2_mechanism} separates the intervention
surfaces the effect does and does not appear on;
\S\ref{sec:causal:controls} isolates the architectural factor against a
Pythia attention baseline and a random-bucketing control; and
\S\ref{sec:causal:niah} transfers the effect to retrieval.

\subsection{BOS coupling at channel granularity (Mamba-1)}
\label{sec:causal:m1_channel}

Ablating the bos-specialist channel set across all layers
(\texttt{gate\_zero} $\times$ bos\_specialist $\times$
\texttt{L\_all}) on Mamba-1 130M produces a differential of
$+3.76$ nats on both wikitext-2 and Pile-10k against the
size-matched random complement, agreeing to two significant
figures ($\Delta < 0.01$ nats). The differential persists at the
larger scale, reaching $+2.13$ on wikitext-2 and $+1.64$ on
Pile-10k at Mamba-1 2.8B (Figure~\ref{fig:selectivity}a), roughly
half the 130M effect, consistent with channel specialty being concentrated
in early-layer state initialisation that scales sub-linearly with
width. Setting the gate parameter to one
(\texttt{gate\_one}) is a dose-response variant that separates
selective gating from static channel activity. On Mamba-1 130M
the \texttt{gate\_one} differential reaches $+26.62$ on wikitext-2
and $+31.53$ on Pile-10k; at 2.8B it is $+19.94$ and $+19.88$,
within $0.06$ nats cross-dataset. It amplifies the
\texttt{gate\_zero} effect by an order of magnitude, consistent
with a coupling that runs through the SSM selectivity
mechanism~\citep{gu2024mamba}. Layer-wise, $46\%$ of the all-layer
effect concentrates at $L_0$ at 130M ($+1.81$ vs $+3.97$ at
\texttt{L\_all}) but only $13\%$ at 2.8B ($+0.25$ vs $+1.87$):
BOS handling localises at the bottom of the smallest network and
redistributes across layers with scale, a $3.5\times$ drop in
$L_0$ localisation consistent with growing redundancy. Reading
this as ``BOS handling broadens with scale'' would be a
task-averaging artefact, since mid-layer ablation is null on
BOS-context at both scales ($+0.054$ / $+0.046$;
Appendix~\ref{sec:appendix:wt_specific}).

\subsection{BOS coupling at head granularity (Mamba-2)}
\label{sec:causal:m2_head}

At head granularity, single-bucket bos-specialty remains positively
causal but with substantially smaller magnitude than at channel
granularity. Two facts complicate reading this set as the circuit.
First, a near-disjoint top-activation probe matches or exceeds its
ablation effect (\S\ref{sec:causal:m2_func_decomp}), so the
localisation is not unique. Second, a separate and larger head set, the dual heads at
$3$--$7\times$ the count, is representationally similar on
boundary classes and carries more aggregate BOS-causal mass under
\texttt{gate\_one} ablation, yet does not share the bos-specialist
causal profile. We establish these in steps: the single-bucket effect
is positively causal but small and survives a $4\times$ context-length
check; the dual set's aggregate dominance is set-size driven; and a
functional decomposition then separates the sets by ablation target and
shows the localisation is non-unique
(\S\ref{sec:causal:m2_func_decomp}).

\paragraph{Single-bucket effect: positive but small.}
\label{sec:causal:m2_30seed}
All 8 cells of the $30$-seed grid (2 scales
$\times$ 2 interventions $\times$ 2 corpora) show positive
single-bucket bos-specialist differentials with bootstrap CI
excluding zero (Table~\ref{tab:m2_30seed}). The widest cell,
M-2 1.3B / \texttt{gate\_one} / Pile, would be marginal under
document-level resampling (\S\ref{sec:limitations}). Cross-dataset
agreement is tight on \texttt{gate\_one}, with $|\Delta| \leq 0.022$
nats on both M-2 scales, and looser on \texttt{gate\_zero}: $0.011$ nats
at 1.3B and $0.42$ nats at 2.7B, where Pile-10k yields the
larger magnitude ($+2.26$ vs.\ $+1.84$) but both CIs still
exclude zero.

\begin{table}[t]
  \centering
  \footnotesize
  \setlength{\tabcolsep}{2.5pt}
  \begin{tabular}{llrrrl}
    \toprule
    Model & Interv & spec & $\bar{x}^{\text{c}}_{30}$ & \textbf{diff} & CI$_{95}$ \\
    \midrule
    \multirow{4}{*}{1.3B}
      & g\_zero wt-2 & 1.60 & 0.48 & $\mathbf{1.12}$ & {\tiny [0.95,1.28]} \\
      & g\_zero Pile & 1.56 & 0.45 & $\mathbf{1.11}$ & {\tiny [0.99,1.22]} \\
      & g\_one  wt-2 & 2.40 & 1.80 & $\mathbf{0.60}$ & {\tiny [0.18,0.93]} \\
      & g\_one  Pile & 2.42 & 1.80 & $\mathbf{0.62}$ & {\tiny [0.16,0.97]} \\
    \midrule
    \multirow{4}{*}{2.7B}
      & g\_zero wt-2 & 2.06 & 0.22 & $\mathbf{1.84}$ & {\tiny [1.74,1.93]} \\
      & g\_zero Pile & 2.39 & 0.13 & $\mathbf{2.26}$ & {\tiny [2.18,2.32]} \\
      & g\_one  wt-2 & 3.14 & 0.66 & $\mathbf{2.48}$ & {\tiny [2.30,2.65]} \\
      & g\_one  Pile & 2.94 & 0.46 & $\mathbf{2.48}$ & {\tiny [2.34,2.61]} \\
    \bottomrule
  \end{tabular}
  \caption{Mamba-2 single-bucket bos-specialist effect under
    $30$ non-adjacent random complement seeds on the BOS-context
    target at \texttt{L\_all}, for wikitext-2 and Pile-10k. Columns: \texttt{spec} = specialist mean
    $\Delta$NLL; $\bar{x}^{\text{c}}_{30}$ = 30-seed
    random-complement mean; \textbf{diff} = $\texttt{spec} -
    \bar{x}^{\text{c}}_{30}$. CI is $5{,}000$-bootstrap on the
    seed mean.}
  \label{tab:m2_30seed}
\end{table}

The M-2 differentials depend on scale relative to the Mamba-1 channel
effect of $+2.13$ at 2.8B (\S\ref{sec:causal:m1_channel}). At 1.3B they
are two to three times smaller, $+0.60$ to $+1.12$; at 2.7B they are
comparable, $+1.84$ to $+2.48$. The head-granularity effect closes the
gap with scale.

\paragraph{Robustness at $4\times$ context length.}
\label{sec:causal:m2_ctx}
At $2048$ ctx, $7$ of $8$ cells reproduce the $512$-ctx
direction with bootstrap CI excluding zero (the one cell that
weakens, M-2 1.3B / \texttt{gate\_one} / Pile-10k, was already
the widest-CI cell at $512$ ctx). The dual-set aggregate effect
(\S\ref{sec:causal:m2_dual}) is robust at both context lengths; full breakdown in
Appendix~\ref{sec:appendix:ctx_2048}.

\paragraph{Dual-head aggregate dominance.}
\label{sec:causal:m2_dual}
At \texttt{gate\_one} ablation on BOS-context at \texttt{L\_all}, the
dual set carries $4.8$--$6.4\times$ the single-bucket differential on
both M-2 scales and both corpora. Both sets are deterministic, so the
ratio carries no seed-driven variance, but it is set-size driven:
across $\tau \in \{0.3, 0.5, 0.7, 1.0\}$ it ranges from $1.02$ to
$6.41\times$ while the per-unit ratio stays near $1\times$, and
\texttt{gate\_zero} and the newline target give different multipliers
(Appendix~\ref{sec:appendix:phenomenology_extra}). The load-bearing
claim is the functional decomposition of
\S\ref{sec:causal:m2_func_decomp}, not the multiplier.

\subsubsection{Functional decomposition by ablation target (F1)}
\label{sec:causal:m2_func_decomp}

The 4.8--$6.4\times$ aggregate ratio is set-size driven. To
separate set size from per-unit causal effect, we re-evaluate
\texttt{gate\_zero} on each head set against both target-token
classes: BOS-context tokens and tokens whose target is a newline. For
each cell (head set, target, scale, corpus) we draw 30
size-matched random complement seeds and report the spec--comp
differential with bootstrap $95\%$ CI (Table~\ref{tab:f1_decomp};
Appendix~\ref{sec:appendix:f1}).

\begin{table}[t]
  \centering
  \footnotesize
  \setlength{\tabcolsep}{3pt}
  \begin{tabular}{ll|cc|cc}
    \toprule
    & & \multicolumn{2}{c|}{M-2 1.3B} & \multicolumn{2}{c}{M-2 2.7B} \\
    \cmidrule(lr){3-4} \cmidrule(lr){5-6}
    set & target & wt-2 & Pile & wt-2 & Pile \\
    \midrule
    bos-spec\textsuperscript{$\dagger$} & wt\_bos & $+1.12$ & $+1.11$ & $+1.84$ & $+2.26$ \\
    dual & wt\_bos & $-0.40$ & $-0.14$ & $+1.14$ & $+1.00$ \\
    bos-spec & wt\_nl & $+3.11$ & $+0.75$ & $+7.42$ & $+1.57$ \\
    dual & wt\_nl & $-1.08$ & $-0.21$ & $-5.79$ & $+0.53$ \\
    \bottomrule
  \end{tabular}
  \caption{F1 functional decomposition: $\Delta$NLL differentials
    (spec mean $-$ size-matched 30-seed random complement mean)
    on wikitext-2 and Pile-10k at $\tau = 0.5$,
    \texttt{gate\_zero}, \texttt{L\_all}. Point estimates; the
    corresponding CIs resample the $30$ complement seeds, as in
    Table~\ref{tab:m2_30seed}, and are listed with per-scope means in
    Appendix~\ref{sec:appendix:f1}.
    \textsuperscript{$\dagger$}Replicated from
    \S\ref{sec:causal:m2_30seed} for direct comparison; comp
    baseline is the 30-seed bank from Table~\ref{tab:m2_30seed}.
    130M cells in Figure~\ref{fig:f1}.}
  \label{tab:f1_decomp}
\end{table}

\paragraph{Non-unique localisation.}
Cosine similarity is weak evidence of function here. Dual heads align
on BOS and newline at $\cos = 0.89$--$0.90$, against $0.53$--$0.54$ for
the specialists. The alignment is not theirs alone. A size- and
depth-matched random null already reaches $0.82$--$0.86$, and
generalist heads reach $0.90$--$0.91$ despite being selected on no
two-class rule at all
(Appendix~\ref{sec:appendix:mechanism_check}). Ablation separates the
sets where cosine does not. The dual set does not reproduce the
bos-specialist newline-target effect of Table~\ref{tab:f1_decomp}, and
the bos-specialist set is in turn not the only set that produces it. A
top-activation probe $B$ that shares just $13$ of $254$ heads with the
single-bucket set $A$, a Jaccard of $0.03$, produces a BOS-context
effect at least as large at both scales, so the single-bucket set is
not the unique locus.

That leaves one way for the two probes to be causally equivalent after
all: the effect might sit entirely in the $11$--$13$ heads they share,
leaving the near-disjoint membership a surface fact about the rest. We
test this by ablating the shared core $A \cap B$ and the two difference
sets on their own (Table~\ref{tab:e2}). In each of the four scale and
target combinations, at least one difference set, containing no shared
head, carries an effect whose paired CI excludes zero. Which side
carries it shifts with scale and target. At 1.3B it is the top-$K$ side
on both targets, $+2.01$ and $+3.53$ nats. At 2.7B on BOS-context it
flips to the single-bucket side, $+0.82$, with the top-$K$ remainder at
$-0.16$ $[-0.32, 0.01]$. On 2.7B newline both sides carry it, $+6.58$
and $+1.68$. The shared core is causally active on its own, $+1.02$ to
$+3.40$ across the four rows, but it does not account for the whole
effect: removing it from either probe leaves a set that is still
load-bearing. Causal support is therefore neither confined to the
intersection nor fixed to one probe's private heads, and which heads
carry it depends on scale and target. The 2.7B BOS-context row is the
sharpest single case, with the top-$K$ effect concentrated in $13$ of
$254$ heads carrying roughly an order of magnitude more effect per head
than the remaining $241$.

\begin{figure}[t]
  \centering
  \includegraphics[width=\textwidth]{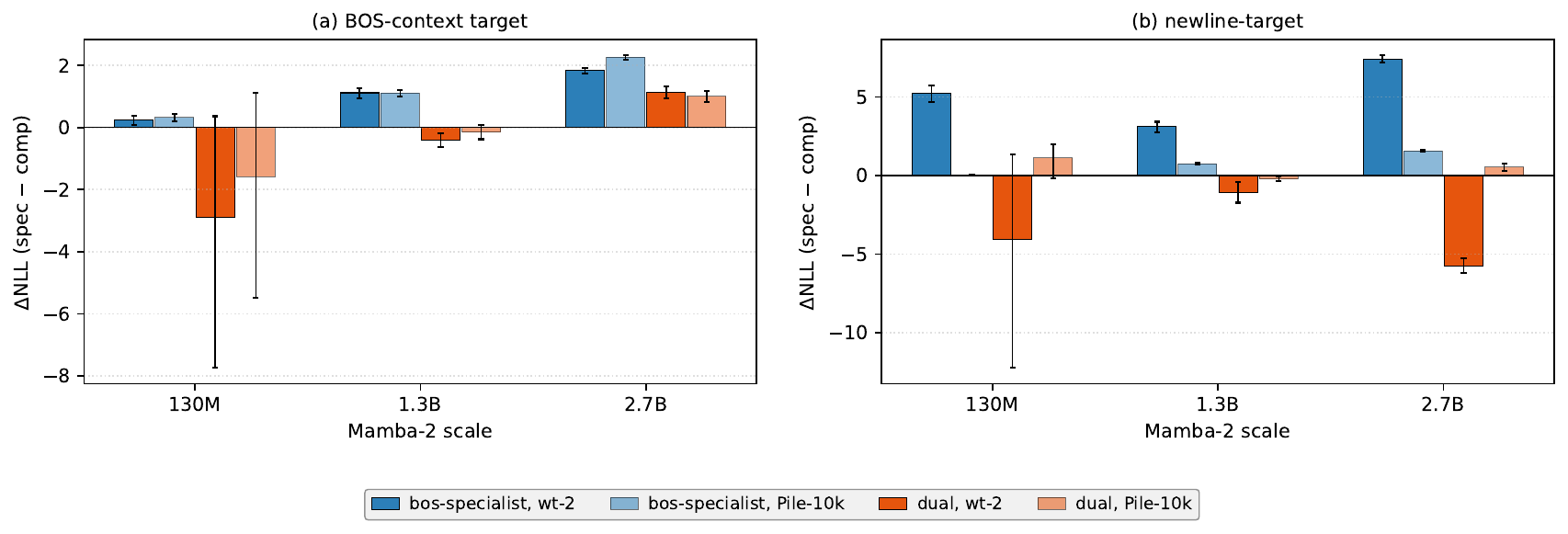}
  \caption{F1 functional decomposition across three M-2 scales:
    spec~$-$~size-matched random complement differentials
    (\texttt{gate\_zero}, \texttt{L\_all}, 30-seed bootstrap
    $95\%$ CI). (a) BOS-context target. (b) Newline-target.
    Per-cell breakdown in \S\ref{sec:causal:m2_func_decomp} and
    Appendix~\ref{sec:appendix:f1}.}
  \label{fig:f1}
\end{figure}

We anchor on BOS-context
because the boundary token is already in context; the same ordering
holds on a newline-prediction target, where the token is not yet
observed (Appendix~\ref{sec:appendix:controls}). The bos-specialist and dual
sets, by contrast, are not interchangeable. Ablating them together on
the 2.7B newline target costs $+9.36$ nats against an additive baseline
of $+6.63$. On BOS-context the same joint ablation is sub-additive, so
the interaction depends on the target
(Appendix~\ref{sec:appendix:controls}). The dual set's effect on
BOS-context prediction also grows with scale, from $-2.89$/$-1.59$ nats
at 130M to $+1.14$/$+1.00$ at 2.7B on wt-2 and Pile. The dual role
separates from BOS-context prediction only in the larger models
(Figure~\ref{fig:f1}).

\begin{table}[t]
  \centering
  \footnotesize
  \setlength{\tabcolsep}{3.5pt}
  \begin{tabular}{ll rr rr}
    \toprule
    & & \multicolumn{2}{c}{full probes} & \multicolumn{2}{c}{shared core removed} \\
    \cmidrule(lr){3-4} \cmidrule(lr){5-6}
    scale & target & $A$ & $B$ & $A \setminus B$ & $B \setminus A$ \\
    \midrule
    \multirow{2}{*}{1.3B}
      & BOS-ctx & $+0.81$ {\tiny [0.55,1.08]} & $+3.22$ {\tiny [2.76,3.66]}
                & $-1.11$ {\tiny [-1.30,-0.91]} & $\mathbf{+2.01}$ {\tiny [1.70,2.31]} \\
      & newline & $+1.76$ {\tiny [1.29,2.25]} & $+8.24$ {\tiny [7.97,8.49]}
                & $-1.29$ {\tiny [-1.49,-1.08]} & $\mathbf{+3.53}$ {\tiny [3.14,3.92]} \\
    \midrule
    \multirow{2}{*}{2.7B}
      & BOS-ctx & $+1.70$ {\tiny [1.30,2.08]} & $+2.40$ {\tiny [2.01,2.79]}
                & $\mathbf{+0.82}$ {\tiny [0.57,1.09]} & $-0.16$ {\tiny [-0.32,0.01]} \\
      & newline & $+6.79$ {\tiny [6.44,7.13]} & $+7.34$ {\tiny [6.82,7.89]}
                & $\mathbf{+6.58}$ {\tiny [6.18,6.97]} & $\mathbf{+1.68}$ {\tiny [1.23,2.14]} \\
    \bottomrule
  \end{tabular}
  \caption{E2 probe split under \texttt{gate\_zero} at \texttt{L\_all}:
    $\Delta$NLL against a size-matched 30-seed random complement,
    paired document bootstrap. $A$ is the single-bucket bos-specialist
    probe, $B$ the top-$K$ probe; $A \cap B$ holds $11$ heads at 1.3B
    and $13$ at 2.7B. Bold marks a difference set whose CI excludes
    zero: at least one appears in every row. Set sizes and the
    $A \cap B$ column are in
    Appendix~\ref{sec:appendix:controls}.}
  \label{tab:e2}
\end{table}

\subsubsection{Intervention-surface dissociation at head granularity}
\label{sec:causal:m2_mechanism}

The probe selects heads by their $\Delta$ activation, so a natural test
is whether perturbing that selectivity reproduces the ablation effect.
On the Mamba-2 bos-specialist heads it does not. We hold the head set and the scored positions fixed and change only the
intervention surface. The output knockout then exceeds either $\Delta$
intervention by $1.40$ and $3.82$ nats at 1.3B, on BOS-context and
newline, and by $1.70$ and $7.17$ nats at 2.7B
(Table~\ref{tab:delta_vs_gate}). The
$\Delta$ interventions themselves move NLL by at most $0.16$ nats. The
set dissociates by \emph{intervention surface}: the $\Delta$ signal the
probe reads carries little of the effect, and the heads' output carries
most of it. This locates the effect on the output surface without
identifying the mediating pathway, since \texttt{gate\_zero} and
\texttt{u\_zero} are identical in every cell and both act as output
knockouts.

The same intervention names do not denote the same operation across the
two architectures. In Mamba-1, \texttt{delta\_zero} zeros the full
post-softplus timestep including the learned bias; in Mamba-2 it zeros
only the pre-bias input, leaving the bias-driven timestep in place. The
Mamba-1 rows are therefore a stronger input-side perturbation, and they
behave accordingly: on BOS-context the bos-specialist channels respond
to every intervention well above a size-matched random complement, with
differentials of $+2.93$ and $+1.74$ nats under \texttt{delta\_zero} at
130M and 2.8B. On the newline target the ordering does not hold, and
the random complement exceeds the specialists in three cells, including
\texttt{delta\_zero} at 2.8B ($+0.03$ against $+0.68$). We read the
Mamba-1 input-side result as confined to the BOS-context endpoint.

\begin{table}[t]
  \centering
  \footnotesize
  \setlength{\tabcolsep}{4pt}
  \begin{tabular}{ll rr rr rr}
    \toprule
    & & \multicolumn{2}{c}{\texttt{delta\_zero}} & \multicolumn{2}{c}{\texttt{delta\_median}} & \multicolumn{2}{c}{output} \\
    \cmidrule(lr){3-4} \cmidrule(lr){5-6} \cmidrule(lr){7-8}
    model & unit & bos & nl & bos & nl & bos & nl \\
    \midrule
    M-1 130M & ch   & $+3.02$ & $+0.93$ & $+0.30$ & $+0.65$ & $+3.97$ & $+1.15$ \\
             &      & {\tiny [2.28,3.77]} & {\tiny [0.51,1.38]} & {\tiny [0.08,0.51]} & {\tiny [0.41,0.88]} & {\tiny [3.30,4.62]} & {\tiny [0.70,1.61]} \\
    M-1 2.8B & ch   & $+1.71$ & $+0.03$ & $+0.39$ & $+3.03$ & $+1.87$ & $+0.62$ \\
             &      & {\tiny [1.38,2.07]} & {\tiny [-0.36,0.43]} & {\tiny [0.21,0.60]} & {\tiny [2.52,3.53]} & {\tiny [1.47,2.27]} & {\tiny [0.20,1.03]} \\
    M-2 1.3B & head & $+0.10$ & $-0.08$ & $+0.09$ & $-0.03$ & $+1.49$ & $+3.74$ \\
             &      & {\tiny [0.06,0.14]} & {\tiny [-0.15,-0.01]} & {\tiny [0.06,0.13]} & {\tiny [-0.09,0.04]} & {\tiny [1.16,1.81]} & {\tiny [3.28,4.23]} \\
    M-2 2.7B & head & $+0.03$ & $+0.16$ & $+0.03$ & $+0.16$ & $+1.73$ & $+7.32$ \\
             &      & {\tiny [-0.02,0.09]} & {\tiny [0.06,0.27]} & {\tiny [-0.02,0.09]} & {\tiny [0.09,0.23]} & {\tiny [1.32,2.14]} & {\tiny [6.64,7.93]} \\
    \bottomrule
  \end{tabular}
  \caption{Input-side (\texttt{delta\_zero}, \texttt{delta\_median})
    against output-side ablation (\texttt{gate\_zero}, identical to
    \texttt{u\_zero}) on the bos-specialist set at \texttt{L\_all},
    wikitext-2 main grid, for BOS-context (bos) and newline (nl)
    targets. Cells are mean $\Delta$NLL under each intervention on the
    same head or channel set, with a $5{,}000$-sample document
    bootstrap $95\%$ CI; the surfaces are compared within a row. At both
    M-2 scales the $\Delta$ and output intervals are disjoint by more
    than a nat. \texttt{delta\_zero} zeros the post-softplus timestep in
    Mamba-1 and the pre-bias input in Mamba-2, so the M-1 and M-2 rows
    are not the same perturbation.}
  \label{tab:delta_vs_gate}
\end{table}

\subsection{Cross-architecture and substrate controls}
\label{sec:causal:controls}

\paragraph{Pythia attention heads.}
\label{sec:causal:pythia}

To separate head granularity per se from architecture-specific
dual allocation, we run the same specialist classification and
ablation pipeline on four Pythia attention-head checkpoints. Pythia heads are classified as
\texttt{bos\_specialist} by mean attention-to-BOS strength above
$0.10$ from positions of each boundary class
\citep{xiao2024streamingllm,gu2024attentionsinks};
this threshold has different semantics from Mamba's pre-softplus
$\Delta$ threshold.
We zero out classified specialist heads across all layers and
score first-eight-tokens-after-BOS NLL on wikitext-2 and Pile-10k,
comparing against 30-seed size-matched non-specialist
complements (same protocol as \S\ref{sec:causal:m2_30seed}).

All four Pythia checkpoints show clean positive selectivity on
both corpora, all eight CIs strictly excluding zero
(cross-dataset agreement $0.05$--$0.30$ nats per cell,
same direction; Appendix~\ref{sec:appendix:pythia_full}). Three
architecture-granularity combinations (Mamba-1 channels,
Mamba-2 heads, Pythia heads) therefore exhibit positive
single-bucket selectivity in the same direction. Mamba-2 head
granularity additionally exhibits dual-head dominance, which
Mamba-1 channels do not. Applying the same multi-class rule (exactly 2 classes above
threshold) to Pythia yields a sparse dual set covering only
$2$--$12\%$ of heads per model, against $27$--$35\%$ on
Mamba-2. Ablation effects sit at the noise level: $\leq 0.16$
nats with the dual-to-bos-spec ratio between $-4.6\times$ and
$0.6\times$ (Appendix~\ref{sec:appendix:pythia_full}), in
contrast to Mamba-2's multi-nat dual differentials.
The Pythia-1.4B differential of $+2.22$ nats is comparable to
Mamba-1 2.8B's $+2.13$ nats; the Pythia drop from $+2.22$ at
1.4B to $+1.58$ at 2.8B parallels the same saturating-with-scale
pattern Mamba-1 channels show ($+3.76 \to +2.13$).

\paragraph{Random channel bucketing (T3).}
\label{sec:causal:t3}
The $5\%$ (Mamba-2 head) vs.\ $29\%$ (Mamba-1 channel)
bos-specialist gap (Table~\ref{tab:bos_fraction}) admits a natural
substrate-capacity hypothesis: that Mamba-2's small per-layer unit
budget ($H/C = 20$ at 2.7B, where $H$ is the per-layer unit count
(heads in M-2, buckets in bucketed M-1 below) and $C = 4$
classes) forces units to multiplex across boundary classes,
suppressing single-bucket specialty and elevating dual specialty. We test
this directly by random channel bucketing of Mamba-1 2.8B. We partition
each layer's $5{,}120$ channels into random groups of $w$ channels and
classify each bucket by its bucket-mean $\Delta\bar{a}$ per class, at
the same $\tau = 0.5$. Sweeping $w$ from $1$ to $64$ brings Mamba-1
down to the per-layer unit budget of M-2 2.7B at the widest setting.

Aggregation alone does not reproduce M-2's distribution. Across the
whole sweep the bucketed bos-specialist fraction stays at
$0.29$--$0.30$ and the dual fraction at $0.28$--$0.30$, against $0.050$
and $0.354$ for M-2 2.7B at the matched $H/C$: coarsening the unit
leaves Mamba-1 channel statistics essentially unchanged
(Appendix~\ref{sec:appendix:t3}). This is consistent with a role for
Mamba-2's head-shared $\Delta$ projection, though head dimension, state
width, and training data are not isolated. A Phase B ablation grid on
bucketed Mamba-1 at $w = 8$ yields sub-nat differentials, an order of
magnitude below M-2 head-level differentials.

\subsection{Long-context retrieval transfer (RULER NIAH)}
\label{sec:causal:niah}

We test whether the bos-specialist set transfers to a
qualitatively different task by running a RULER-style
Needle-in-a-Haystack~\citep{hsieh2024ruler} probe at $1024$ and $2048$
context on the two flagship Mamba checkpoints. We report a thresholded
teacher-forced success rate: a trial counts as retrieved when the mean
answer-token NLL is below $3.0$, with $10$ trials at each of three
needle depths ($30$ per condition). At $1024$ ctx, all three
conditions separate cleanly: the unablated model retrieves in all $30$
trials, bos-specialist ablation in none, and a single size-matched
random complement in all $30$. That complement is one random draw. It rules out a generic
size-matched-ablation artefact, but it is not a control distribution.
The stronger control is a magnitude-matched non-specialist set: per
layer, the top-$K$ non-bos-specialist heads ranked by mean
$|\Delta_{\text{pre}}|$ across all tokens, with $K$ matching the
bos-specialist count.
The two architectures answer that control differently. At Mamba-2 2.7B
the magnitude-matched complement leaves retrieval intact at both
context lengths, $1.00$ and $0.97$, so what drives the collapse is
BOS-specialty and not raw activation magnitude. At Mamba-1 2.8B the
same complement collapses retrieval outright, $0.00$ at both lengths,
which fits the broader channel-granularity substrate of
\S\ref{sec:causal:m1_channel}: there, several different subsets are
causally important. The per-token NLL gap on
the answer span under bos-specialist ablation is $\sim\!8.5$ nats
on M-1 2.8B and $\sim\!9.7$ nats on M-2 2.7B, substantially
larger than the corresponding BOS-context differentials. At $2048$ ctx the separation holds on M-2 2.7B (baseline $1.00$,
bos-specialist ablation $0.00$, complement $1.00$); M-1 2.8B, trained
at $2048$, shows only mild baseline degradation at mid-depth. The
Mamba-2 dual ablation produces partial degradation at both lengths
(mean success rate $0.40$ / $0.27$), converging with the
$4.8$--$6.4\times$ aggregate ratio (\S\ref{sec:causal:m2_head}):
bos-specialist and dual heads have overlapping but non-identical
retrieval substrates. Beyond $2048$ ctx both checkpoints' baseline
retrieval collapses (both pretrained at
$2048$~\citep{ye2025longmamba}), so longer-context intervention lacks
a meaningful baseline. Full breakdowns and the long-context compute
constraint are in Appendix~\ref{sec:appendix:niah_breakdown}
and~\ref{sec:appendix:niah_compute}.

\section{Discussion}
\label{sec:discussion}

\paragraph{What the evidence supports, and how strongly.}
The three findings rest on different kinds of evidence, and we mark the
difference. \emph{Causal non-uniqueness} is established by ablation:
two near-disjoint probes are each load-bearing, and with the shared
core removed at least one remainder still carries an effect whose
paired CI excludes zero in every scale and target combination
(\S\ref{sec:causal:m2_func_decomp}). For \emph{representation and
function} the ablation is the direct evidence: the dual set fails to
reproduce the bos-specialist profile under class-conditional ablation.
The cosine gap is the weaker half, since a matched null and the
generalist set both recover most of it.
\emph{Intervention-surface dissociation} is established at the level of
surfaces: the $\Delta$ and output intervals are disjoint by more than a
nat at both M-2 scales, which places the effect on the output surface.
Identifying the mediating pathway would take an intervention that
separates the gate from the head output, which \texttt{gate\_zero} and
\texttt{u\_zero} do not (\S\ref{sec:causal:m2_mechanism}). A
representational signature at Mamba-2 head granularity thus marks a
behaviourally important head set without pinning down the circuit.

\paragraph{How often should this be expected?}
The failure is not uniform across the models we test, and its
distribution is what makes it actionable. It does not appear at Mamba-1
channel granularity, where $\Delta$ is computed per channel and the
probe's units carry the causal effect, and it does not appear in the
four Pythia checkpoints, where the multi-class structure is absent
altogether. It appears at Mamba-2 head granularity, where one $\Delta$
gate is shared across a head's channels. That contrast points to the factor. When a model gates more coarsely
than it computes, a single boundary class no longer has a unit of its
own, and the causal work spreads across overlapping sets that any one
probe recovers only in part. The prediction is testable and cheap to check on a new
architecture: substrates that gate more coarsely than they compute
should show the dissociation, and substrates with independent per-unit
gating should not. Our controls are consistent with this without
isolating it, since head dimension, state width, and training data
differ alongside the gating.

\paragraph{Implications.}
Single-signal readings of attention
sinks~\citep{xiao2024streamingllm,gu2024attentionsinks} and Mamba
interpretability~\citep{chiang2025quamba,ye2025longmamba} suffice
at fine granularity but are incomplete at Mamba-2 heads; prior
single-bucket localisations on head-shared substrates~\citep{sharma2024locating,ali2024hidden}
warrant re-audit under multi-class aggregation. The contrast with
attention sinks runs deeper than the measurement space. SDPA sink heads
are \emph{replaceable}~\citep{qiu2025gated}, and their value norms are
suppressed~\citep{xiao2024streamingllm}. These heads are neither: they
carry causal mass for BOS-context and for retrieval
(\S\ref{sec:causal:m1_channel}, \S\ref{sec:causal:niah}). The Mamba-2
state sink is \emph{distributed} across overlapping head sets that no
single probe recovers in full, and those sets do real work. What follows from this is a check, not a new method. Take a
localisation read off one probe on a head-shared substrate. Run a
second probe of a different form, then ablate the two difference sets.
That tells you whether the located set is \emph{a} circuit or
\emph{the} circuit. In our grid the check costs two ablation runs, and
it changes the answer.

\section{Limitations}
\label{sec:limitations}

\paragraph{Datasets and models.}
The intervention grid covers four Mamba checkpoints (Mamba-1
130M / 2.8B, Mamba-2 1.3B / 2.7B) on
wikitext-2~\citep{merity2017wikitext} and
Pile-10k~\citep{gao2020pile}; the 30-seed cross-dataset
confirmation covers all 8 M-2 head-granularity headline cells.
Mamba-2 130M is added for the F1 scale ladder
(\S\ref{sec:causal:m2_func_decomp}) and the phenomenology
cross-method correlations only.
Three further wikitext-2-specific findings do not replicate on
Pile-10k (Appendix~\ref{sec:appendix:wt_specific}).
Pythia~\citep{biderman2023pythia} 160M--2.8B is the
cross-architecture baseline; hybrid SSM--attention models such
as Zamba~\citep{glorioso2024zamba} are out of scope. Primary
scoring is first-eight-tokens-after-boundary NLL on
BOS-context / newline plus RULER NIAH retrieval at
$1024$ and $2048$ ctx; formal-task evaluations (parity, Dyck-2,
$S_3$-permutation) are deferred.
Untested classes, including punctuation and
\texttt{whitespace\_word}, may carry additional causal sets.

\paragraph{Effect size.}
Mamba-2 single-bucket differentials are small ($+0.25$--$+2.48$
on BOS-context vs.\ $+3.76$ for M-1 130M); the M-2 1.3B
\texttt{gate\_one} CI is wide but excludes zero. At M-2 130M the
dual set is small enough that its ablation CIs are wide (point
estimates remain monotonic with scale; \S\ref{sec:causal:m2_func_decomp}).
T3 rules out substrate granularity alone, not head dimension,
state width, or training-data alternatives.

\paragraph{Uncertainty.}
The headline bootstrap CIs resample the $30$ complement seeds and
condition on the fixed probe-selected set; they do not propagate
document-sampling or specialist-selection uncertainty. A document-level
resampling would widen the marginal cells, and full set-selection
uncertainty is left to future work. The NIAH complement is a single
random draw, not a multi-seed control.

\section*{Ethics Statement}
\label{sec:ethics}

\paragraph{Data and models.} All experiments use publicly
released language models (Mamba-1
\citep{gu2024mamba}, Mamba-2 \citep{dao2024mamba2},
Pythia~\citep{biderman2023pythia}) under their respective public
release terms, and publicly released
text corpora (wikitext-2~\citep{merity2017wikitext},
Pile-10k~\citep{gao2020pile}). No human subjects, no personally
identifying information, and no proprietary data are involved.
We comply with the data release terms on attribution and
redistribution.

\paragraph{Compute and environmental impact.} Summing the logged
per-cell wall-clock over the intervention sweeps reported here gives
approximately $170$ GPU-hours. Pilot runs, specialty extraction, and
the retrieval-transfer runs are not separately timed and are
additional. All but one experiment runs on a single RTX 3090
24\,GB; the $2048$-context Mamba-2 2.7B cells exceed $24$\,GB
and were run on an H800 80\,GB card
(Appendix~\ref{sec:appendix:niah_compute}). No model is trained or
fine-tuned, so the total is small next to a pre-training run. We cache
specialty classifications and reuse the same chunk preparation across
cells to avoid repeated forward passes.

\paragraph{Reproducibility.} The grid is pre-registered with an
amendment trail documenting design changes between v0 and v6
(Appendix~\ref{sec:appendix:prereg}).
Random complement protocols use stratified non-adjacent seeds
(master seed $13$, gap $\geq 61$ across $[0, 10\,000)$),
$5\,000$-sample bootstrap on the seed mean, and deterministic
specialty classification.

\paragraph{Potential applications and risks.} Interpretability of
selective state-space language models has primarily safety- and
transparency-positive applications: it enables informed head-level
diagnosis, principled pruning, and audit tooling for production
systems. Misuse to selectively damage downstream capability would
require white-box access to the deployed model, which is the same
threat surface as standard model deployment. We are not aware of
risks beyond those already present in publishing model weights
and standard interpretability analyses.

\bibliographystyle{tmlr}
\bibliography{refs}

\appendix
\section{Protocol and Reproducibility}

\subsection{Pre-registration}
\label{sec:appendix:prereg}

The intervention grid and its primary predictions were
pre-registered before the grid was run. The Pythia
cross-architecture baseline and the RULER retrieval transfer were
added as timestamped amendments before running. The E1--E4 controls
and the $\Delta$-versus-gate mechanism comparison
(\S\ref{sec:causal:m2_mechanism}) are post-hoc analyses added during
revision.

\subsection{Cross-dataset replication protocol}
\label{sec:appendix:cross_dataset}

Cross-dataset replication uses the
\texttt{NeelNanda/\allowbreak{}pile-10k} subset of The
Pile~\citep{gao2020pile}: $30$ documents per cell, filtered to
contain at least two newline tokens per chunk for parity with the
wikitext-2 sampling. Seeds, bootstrap CI methodology
($n_{\text{boot}} = 5{,}000$), and per-token NLL
scoring on first-eight-tokens-after-BOS are identical to the
wikitext-2 grid. We replicated 102 cells targeting the headline
mechanisms: 24 BOS-coupling cells (\S\ref{sec:causal:m1_channel},
\S\ref{sec:causal:m2_head}), 12 \texttt{gate\_one} selectivity cells
(\S\ref{sec:causal:m2_head}), 12 dual-head dominance cells, 6
cross-task swap cells, 16 layer-locus / mid-layer cells,
and 32 endpoint cells from the newline mirror-symmetry analysis
that ultimately demoted to wikitext-2-specific
(Appendix~\ref{sec:appendix:wt_specific}).

\section{Phenomenology: Cross-Method Agreement and Bimodality}

\subsection{Cross-method tables and per-model breakdown}
\label{sec:appendix:phenomenology_extra}

\suppressfloats[t]

The main body \S\ref{sec:phenomenology:sinkmass} reports the
headline LongMamba $\times$ $\Delta_{\text{pre}}$-right-tail
Spearman correlation result; the per-model values, the parallel
Quamba $\times$ A-log polarisation pair, and the cross-method
agreement breakdown with probe-derived specialist sets are
collected here. Two prior-work metrics
\citep{chiang2025quamba,ye2025longmamba} and two analytic
baselines (right-tail $\Pr(\Delta_{\text{pre}} > 0.7)$ and
$-\log A$ from the stored $A_{\log}$ parameter) are evaluated per
channel/head per layer for all $7$ Mamba checkpoints.

\begin{table}[t]
  \centering
  \small
  \begin{tabular}{lrl}
    \toprule
    Model & $\rho_{L \times \Delta}$ & Verdict \\
    \midrule
    Mamba-1 130M & $+0.198$ & FAIL \\
    Mamba-1 370M & $+0.182$ & FAIL \\
    Mamba-1 1.4B & $-0.023$ & WEAK \\
    Mamba-1 2.8B & $+0.085$ & FAIL \\
    Mamba-2 130M & $-0.282$ & WEAK \\
    Mamba-2 1.3B & $\mathbf{-0.443}$ & \textbf{PASS} \\
    Mamba-2 2.7B & $\mathbf{-0.437}$ & \textbf{PASS} \\
    \bottomrule
  \end{tabular}
  \caption{LongMamba $\times$ $\Delta_{\text{pre}}$-right-tail
    Spearman $\rho$ averaged across all layers per model;
    pre-registered prediction $\rho < -0.3$
    (\S\ref{sec:phenomenology:sinkmass}).}
  \label{tab:m3_la_pred}
\end{table}

\paragraph{Cross-method agreement with probe specialty labels.}
The pre-registered prediction was that $\geq 3$ of $4$ sink-mass
metrics should agree with probe-derived bos-specialist sets at
top-10\% Jaccard $> 0.4$. Observed Jaccard is in $[0.07, 0.15]$
across all 7 models and all 4 lines, substantially above chance
($\sim 5\%$) but well below the $0.4$ prediction. Best agreement
on Mamba-2: LongMamba and A-log polarisation
(Jaccard $\sim 0.14$--$0.15$); worst: $\Delta_{\text{pre}}$
right-tail (Jaccard $\sim 0.03$--$0.07$). Prior sink-mass methods
capture partially complementary signals rather than a single
canonical sink-mass quantity.

\paragraph{Quamba $\times$ A-log polarisation systematic
anti-correlation on Mamba-1.} Quamba (activation outlier) and
A-log polarisation (architectural decay) are consistently
negatively correlated on all four Mamba-1 scales
($\rho \in [-0.25, -0.37]$) but uncorrelated or weakly positive
on Mamba-2 ($\rho \in [-0.01, +0.19]$). At channel granularity,
activation-based and spectral-based sink scores capture distinct
aspects of channel behaviour; at head granularity, the
distinction collapses.

\begin{figure}[t]
  \centering
  \includegraphics[width=0.92\textwidth]{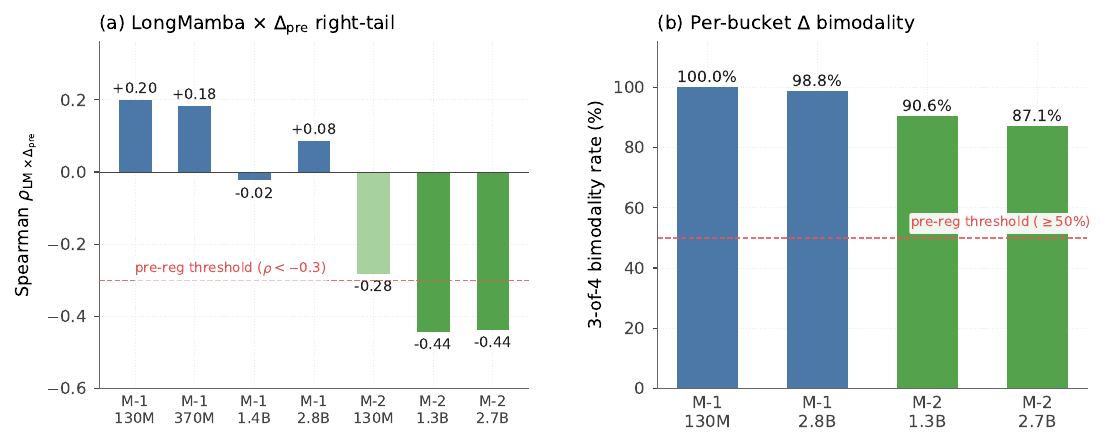}
  \caption{Cross-method phenomenology evidence.
    (a) LongMamba--$\Delta_{\text{pre}}$-right-tail Spearman
    $\rho$ averaged across layers. (b) Per-bucket
    pre-$\Delta$ distribution bimodality rate under the 3-of-4
    method conjunction (Hartigan dip, Silverman peaks,
    GMM $\Delta$BIC, KDE local modes). Findings discussed in
    \S\ref{sec:phenomenology:sinkmass}.}
  \label{fig:phenomenology}
\end{figure}

\paragraph{Per-bucket bimodality 3-of-4 conjunction
(Table~\ref{tab:m4_bimodality}).}
\begin{table}[t]
  \centering
  \footnotesize
  \begin{tabular}{lrr}
    \toprule
    Model & 3-of-4 / total & rate \\
    \midrule
    Mamba-1 130M & 96 / 96   & $\mathbf{100.0\%}$ \\
    Mamba-1 2.8B & 253 / 256 & $\mathbf{98.8\%}$ \\
    Mamba-2 1.3B & 174 / 192 & $\mathbf{90.6\%}$ \\
    Mamba-2 2.7B & 223 / 256 & $\mathbf{87.1\%}$ \\
    \bottomrule
  \end{tabular}
  \caption{Per-bucket pre-$\Delta$ bimodality rate under the
    3-of-4 method conjunction
    (\S\ref{sec:phenomenology:sinkmass}).}
  \label{tab:m4_bimodality}
\end{table}

\paragraph{Threshold sensitivity of the dual-dominance ratio.}
We re-ran the \texttt{gate\_one} $\times$ BOS-context $\times$
\texttt{L\_all} ablation at three alternative thresholds
$\tau \in \{0.3, 0.7, 1.0\}$ (in addition to the paper default
$\tau = 0.5$). At each $\tau$, we re-classified (layer, head)
units from cached per-bucket mean differentials, then ran the
ablation on the new bos-specialist and dual sets. The aggregate
ratio is non-monotonic in $\tau$ and peaks near the default
$\tau = 0.5$ (Table~\ref{tab:tau_sens}); it ranges across
$1.02\times$--$6.41\times$ depending on $\tau$ and cell. The
$|D|/|S|$ set-size dominance is robust across all sampled $\tau$
($\geq 2.16\times$), and the per-unit ratio is consistently
$\leq 1.5\times$. The headline aggregate ratio
$4.8$--$6.4\times$ should therefore be read as the value at the
pre-registered default threshold, not as a $\tau$-invariant
claim; the load-bearing mechanistic claim is the qualitative
BOS--newline alignment documented in
\S\ref{sec:appendix:mechanism_check}.

\begin{table}[t]
  \centering
  \footnotesize
  \setlength{\tabcolsep}{2pt}
  \begin{tabular}{lcccccc}
    \toprule
    & & & \multicolumn{2}{c}{wt-2} & \multicolumn{2}{c}{Pile} \\
    \cmidrule(lr){4-5} \cmidrule(lr){6-7}
    $\tau$ & $|S|$ & $|D|$ & spec & dual & spec & dual \\
    \midrule
    \multicolumn{7}{l}{\emph{M-2 1.3B (ratio under each (spec, dual) pair)}} \\
    $0.3$  & $142$ & $698$  & $+1.42$ & $+6.04$ & $+1.18$ & $+7.53$ \\
    & & & \multicolumn{2}{c}{$4.26\times$} & \multicolumn{2}{c}{$6.36\times$} \\
    $0.5$* & $251$ & $826$  & $+2.13$ & $+10.33$ & $+2.42$ & $+11.54$ \\
    & & & \multicolumn{2}{c}{$\mathbf{4.84\times}$} & \multicolumn{2}{c}{$\mathbf{4.77\times}$} \\
    $0.7$  & $351$ & $860$  & $+2.55$ & $+7.31$ & $+3.31$ & $+7.61$ \\
    & & & \multicolumn{2}{c}{$2.87\times$} & \multicolumn{2}{c}{$2.30\times$} \\
    $1.0$  & $434$ & $939$  & $+4.29$ & $+4.56$ & $+5.30$ & $+5.41$ \\
    & & & \multicolumn{2}{c}{$1.06\times$} & \multicolumn{2}{c}{$1.02\times$} \\
    \midrule
    \multicolumn{7}{l}{\emph{M-2 2.7B}} \\
    $0.3$  & $154$ & $1457$ & $+1.77$ & $+6.60$ & $+1.98$ & $+6.81$ \\
    & & & \multicolumn{2}{c}{$3.72\times$} & \multicolumn{2}{c}{$3.43\times$} \\
    $0.5$* & $254$ & $1810$ & $+2.66$ & $+17.07$ & $+2.94$ & $+16.74$ \\
    & & & \multicolumn{2}{c}{$\mathbf{6.41\times}$} & \multicolumn{2}{c}{$\mathbf{5.70\times}$} \\
    $0.7$  & $370$ & $1886$ & $+3.56$ & $+12.37$ & $+3.76$ & $+12.24$ \\
    & & & \multicolumn{2}{c}{$3.48\times$} & \multicolumn{2}{c}{$3.25\times$} \\
    $1.0$  & $514$ & $1843$ & $+4.64$ & $+10.87$ & $+4.78$ & $+11.39$ \\
    & & & \multicolumn{2}{c}{$2.34\times$} & \multicolumn{2}{c}{$2.38\times$} \\
    \bottomrule
  \end{tabular}
  \caption{Threshold sensitivity of the dual / bos-specialist
    aggregate ratio under \texttt{gate\_one} ablation at
    \texttt{L\_all} on BOS-context. $\tau = 0.5^{*}$ is the
    pre-registered default; discussion of the
    $\tau$-dependence and the set-size dominance
    $|D|/|S| \geq 2.16\times$ in
    \S\ref{sec:appendix:phenomenology_extra}.}
  \label{tab:tau_sens}
\end{table}

\subsection{Bimodality detection per method and strict conjunction}
\label{sec:appendix:bimodality_calibration}

For each (layer, bucket) cell of the four Mamba models tested in
\S\ref{sec:phenomenology:sinkmass}, we record per-method positive
rates and both the pre-registered 3-of-4 conjunction and the strict
4-of-4 conjunction (Table~\ref{tab:bimodality_per_method}).

\begin{table}[t]
  \centering
  \small
  \setlength{\tabcolsep}{4pt}
  \begin{tabular}{lrrrr}
    \toprule
    Model & silver & $\Delta$BIC & 3-of-4 & 4-of-4 \\
    \midrule
    M-1 130M & $100.0$ & $96.9$ & $100.0$ & $96.9$ \\
    M-1 2.8B & $97.7$  & $96.1$ & $98.8$  & $94.9$ \\
    M-2 1.3B & $85.9$  & $58.3$ & $90.6$  & $53.6$ \\
    M-2 2.7B & $81.6$  & $47.3$ & $87.1$  & $41.8$ \\
    \bottomrule
  \end{tabular}
  \caption{Bimodality positive rates (\%) per model. Hartigan
    dip~$\geq 0.05$ and KDE local-mode count~$\geq 2$ both saturate
    at $100\%$ across all four models and are omitted from the
    table; Silverman peaks~$\geq 2$ and GMM $\Delta$BIC~$\geq 10$
    are the discriminating criteria, with the parametric
    $\Delta$BIC the most restrictive on Mamba-2 (lowest rate
    $47.3\%$ on M-2 2.7B). The strict 4-of-4 conjunction drops
    below the $50\%$ pre-registration threshold on Mamba-2 2.7B
    ($41.8\%$); the 3-of-4 conjunction (specified pre-registration
    to retain power against the unimodal null when one method is
    borderline) exceeds $87\%$ on all four models.}
  \label{tab:bimodality_per_method}
\end{table}

\section{Causal Controls and Decompositions}

\subsection{Probe split, threshold sweep, and complement reconciliation}
\label{sec:appendix:controls}

The probe split (E2) and the joint / marginal decomposition (E4) are
summarised in the main text (Tables~\ref{tab:e2} and~\ref{tab:e4}); the
cosine null (E1) is in Appendix~\ref{sec:appendix:mechanism_check}. This
appendix gives the full E2 numbers, the threshold sweep, and the
complement-definition sensitivity.

\paragraph{E2 on the newline-prediction target.}
The main text anchors on BOS-context, where the boundary token is
already in context. On the newline-prediction target, where the token is
not yet observed, the ordering is the same: the single-bucket
\texttt{gate\_zero} effect is $+3.92$ at 1.3B and $+7.50$ at 2.7B,
against a top-$K$ effect of $+8.50$ and $+7.49$.

\paragraph{E2 probe split: full numbers.}
Table~\ref{tab:e2_split_full} gives the absolute and complement-referenced
effects behind Table~\ref{tab:e2}, together with the set sizes. The
single-bucket probe $A$ and the top-$K$ probe $B$ are matched in size by
construction ($251$ heads at 1.3B, $254$ at 2.7B); their intersection
holds $11$ and $13$ heads, so each difference set holds $240$ and $241$.
Absolute columns are the mean $\Delta$NLL against the unablated model;
\textbf{diff} is against a size-matched $30$-seed random complement,
with a paired document bootstrap ($n_{\text{boot}} = 10{,}000$) over the
$30$ scored documents. Pairing the specialist and complement scores
within document removes document difficulty from the interval.

\begin{table}[t]
  \centering
  \footnotesize
  \setlength{\tabcolsep}{3pt}
  \begin{tabular}{ll rr rr}
    \toprule
    & & \multicolumn{2}{c}{absolute} & \multicolumn{2}{c}{diff vs.\ complement} \\
    \cmidrule(lr){3-4} \cmidrule(lr){5-6}
    set (size) & target & 1.3B & 2.7B & 1.3B & 2.7B \\
    \midrule
    \multirow{2}{*}{$A$ (251 / 254)}
      & BOS-ctx & $+1.60$ & $+2.06$ & $+0.81$ {\tiny [0.55,1.08]} & $+1.70$ {\tiny [1.30,2.08]} \\
      & newline & $+3.92$ & $+7.50$ & $+1.76$ {\tiny [1.29,2.25]} & $+6.79$ {\tiny [6.44,7.13]} \\
    \midrule
    \multirow{2}{*}{$B$ (251 / 254)}
      & BOS-ctx & $+3.44$ & $+2.52$ & $+3.22$ {\tiny [2.76,3.66]} & $+2.40$ {\tiny [2.01,2.79]} \\
      & newline & $+8.50$ & $+7.49$ & $+8.24$ {\tiny [7.97,8.49]} & $+7.34$ {\tiny [6.82,7.89]} \\
    \midrule
    \multirow{2}{*}{$A \cap B$ (11 / 13)}
      & BOS-ctx & $+1.03$ & $+1.27$ & $+1.02$ {\tiny [0.63,1.42]} & $+1.26$ {\tiny [0.96,1.57]} \\
      & newline & $+1.18$ & $+3.38$ & $+1.21$ {\tiny [0.89,1.57]} & $+3.40$ {\tiny [3.08,3.76]} \\
    \midrule
    \multirow{2}{*}{$A \setminus B$ (240 / 241)}
      & BOS-ctx & $+0.08$ & $+1.27$ & $-1.11$ {\tiny [-1.30,-0.91]} & $+0.82$ {\tiny [0.57,1.09]} \\
      & newline & $+0.45$ & $+7.42$ & $-1.29$ {\tiny [-1.49,-1.08]} & $+6.58$ {\tiny [6.18,6.97]} \\
    \midrule
    \multirow{2}{*}{$B \setminus A$ (240 / 241)}
      & BOS-ctx & $+2.50$ & $+0.32$ & $+2.01$ {\tiny [1.70,2.31]} & $-0.16$ {\tiny [-0.32,0.01]} \\
      & newline & $+4.25$ & $+2.53$ & $+3.53$ {\tiny [3.14,3.92]} & $+1.68$ {\tiny [1.23,2.14]} \\
    \bottomrule
  \end{tabular}
  \caption{E2 probe split, \texttt{gate\_zero} at \texttt{L\_all} on
    WikiText-2. Sizes are given as 1.3B / 2.7B. The only difference-set
    interval containing zero is $B \setminus A$ on 2.7B BOS-context.}
  \label{tab:e2_split_full}
\end{table}

\paragraph{Joint and marginal ablation (E4).}
Ablating the bos-specialist set $A$ and the dual set $B$ together tests
whether they are redundant. At 2.7B they are not
(Table~\ref{tab:e4}): on the newline target the joint ablation costs
more than the sum of the separate ones, while on BOS-context it costs
less. Both sets are deterministic, so these are point estimates with no
set-sampling variance.

\begin{table}[t]
  \centering
  \footnotesize
  \setlength{\tabcolsep}{4pt}
  \begin{tabular}{llrrrr}
    \toprule
    scale & target & A & B & AB & $A{+}B$ \\
    \midrule
    \multirow{2}{*}{2.7B}
      & newline & $+7.50$ & $-0.87$ & $\mathbf{+9.36}$ & $+6.63$ \\
      & BOS     & $+2.06$ & $+2.75$ & $+4.16$          & $+4.81$ \\
    \bottomrule
  \end{tabular}
  \caption{E4: \texttt{gate\_zero} joint / marginal decomposition at
    2.7B. A $=$ bos-specialist, B $=$ dual, AB $=$ their union,
    $A{+}B$ the additive baseline.}
  \label{tab:e4}
\end{table}

\paragraph{Threshold sweep (E3).}
Sweeping the specialty threshold
$\tau \in \{0.3, 0.4, 0.5, 0.6, 0.7\}$, we report the bos-specialist
spec$-$complement differential on the newline target
(Table~\ref{tab:e3}). At 2.7B the differential is positive at every
$\tau$ on both corpora. At 1.3B it is positive except at wt-2
$\tau = 0.4$, where an anomalous negative absolute bos-specialist
effect drives it to $-1.70$; the sign is otherwise stable.

\begin{table}[t]
  \centering
  \footnotesize
  \setlength{\tabcolsep}{4pt}
  \begin{tabular}{lrrrrr}
    \toprule
    scale / corpus & $\tau.3$ & $.4$ & $.5$ & $.6$ & $.7$ \\
    \midrule
    2.7B wt-2 & $+5.26$ & $+6.53$ & $+7.42$ & $+8.17$ & $+7.92$ \\
    2.7B Pile & $+3.39$ & $+1.20$ & $+1.57$ & $+1.70$ & $+1.19$ \\
    1.3B wt-2 & $+0.40$ & $\mathbf{-1.70}$ & $+3.11$ & $+4.04$ & $+3.99$ \\
    1.3B Pile & $+0.09$ & $+0.90$ & $+0.75$ & $+1.21$ & $+1.84$ \\
    \bottomrule
  \end{tabular}
  \caption{E3: bos-specialist spec$-$complement differential (nats) on
    the newline target across the classification threshold $\tau$;
    $\tau = 0.5$ is the pre-registered main-text value.}
  \label{tab:e3}
\end{table}

\paragraph{Complement definition.}
The dual newline differentials in Table~\ref{tab:f1_decomp} depend on
how the random complement is drawn. The \texttt{complement\_dual} bank
forbids only dual heads, so it can draw bos-specialist heads and widen
the gap. Under a symmetric complement that also forbids bos-specialists,
the dual--newline differential moves from $-1.08$ to $+0.11$ at 1.3B and
from $-5.79$ to $-6.02$ at 2.7B. The 1.3B dual deficit is thus
complement-dependent; the 2.7B result is unchanged.

\subsection{F1 functional decomposition details}
\label{sec:appendix:f1}

The F1 grid (\S\ref{sec:causal:m2_func_decomp}) runs
\texttt{gate\_zero}~$\times$~\texttt{L\_all} on Mamba-2 1.3B and
2.7B against the cross product of head sets
\{bos-specialist, dual, complement\_bos, complement\_dual\} and
targets \{BOS-context, newline\} on wikitext-2 and
Pile-10k. Spec sets are deterministic; complement sets are
size-matched with $30$ non-adjacent stratified seeds (master
$13$). Each cell is bootstrapped with $5{,}000$ replicates on
the per-document $\Delta$NLL distribution
($n_{\text{docs}} = 30$, max-len $512$). Table~\ref{tab:f1_decomp}
in the body reports headline differentials (spec mean minus
seed-averaged complement mean) with $95\%$ bootstrap CI; the
full per-scope absolute $\Delta$NLL means and per-seed bootstrap
CIs are in the released
\texttt{f1\_dual\_function\_test} artefact, one summary per
(scale, corpus) cell. The dual~$\times$
wt\_newline differential at M-2 2.7B wt-2 ($-5.79$ nats) has a
$95\%$ CI of $[-6.25, -5.33]$, well-excluding zero in the
negative direction; the corresponding M-2 1.3B wt-2 differential
($-1.08$) has CI $[-1.71, -0.46]$; M-2 1.3B Pile-10k differential
($-0.21$) has CI $[-0.38, -0.08]$.

\subsection{Dual-head BOS--newline alignment (\texorpdfstring{$\Delta_{\text{pre}}$}{Delta-pre} test)}
\label{sec:appendix:mechanism_check}

We test the boundary-reset hypothesis (\S\ref{sec:discussion}) by
comparing class-conditional cosine similarity on the
$\Delta_{\text{pre}}$ activations of (a) Mamba-2 dual heads vs.\
(b) Mamba-2 single-bucket bos-specialists, on $100$ wikitext-2
documents at $512$ ctx. For each (layer, head) in the head set,
we compute the mean $\Delta_{\text{pre}}$ conditioned on the
next-token class $\in$ \{BOS, newline, punctuation,
whitespace-word, other-word\}; stack head-means into a
class-vector and take pairwise cosine.

\begin{table}[t]
  \centering
  \footnotesize
  \setlength{\tabcolsep}{3pt}
  \begin{tabular}{llrr}
    \toprule
    Model & class pair & dual & bos\_spec \\
    \midrule
    \multirow{4}{*}{M-2 1.3B}
      & cos(BOS, newline)    & $\mathbf{+0.90}$ & $+0.53$ \\
      & cos(BOS, punct.)     & $+0.70$          & $+0.72$ \\
      & cos(newline, punct.) & $+0.73$          & $+0.68$ \\
      & cos(BOS, other-word) & $+0.77$          & $+0.76$ \\
    \midrule
    \multirow{4}{*}{M-2 2.7B}
      & cos(BOS, newline)    & $\mathbf{+0.89}$ & $+0.54$ \\
      & cos(BOS, punct.)     & $+0.47$          & $+0.45$ \\
      & cos(newline, punct.) & $+0.57$          & $+0.85$ \\
      & cos(BOS, other-word) & $+0.66$          & $+0.59$ \\
    \bottomrule
  \end{tabular}
  \caption{Class-conditional $\Delta_{\text{pre}}$ cosine
    similarity on Mamba-2 dual vs.\ bos-specialist head sets,
    $100$ wt-2 docs at $512$ ctx. The BOS--newline row is the
    discriminating signal; other class pairs are similar across
    head sets because $\Delta_{\text{pre}}$ values are
    predominantly positive in $L_2$ space. Full $5 \times 5$
    matrix in \texttt{results/dual\_head\_mechanism/}.}
  \label{tab:dual_head_cosine}
\end{table}

\paragraph{Size- and depth-matched null (E1).}
The dual set is $3$--$7\times$ larger than the bos-specialist set, so
averaging over more units could smooth the class-mean estimate and
inflate the cosine. It does not: subsampling the dual set down to the
bos-specialist size over $K = 200$ random draws gives a mean of
$0.9000$ at 1.3B and $0.8879$ at 2.7B, matching the full-set values of
$0.9004$ and $0.8880$ and well above the single-bucket cosines of
$0.53$--$0.54$. Set size is therefore not what produces the alignment.

Two further comparators show that the alignment is not specific to the
dual set either (Table~\ref{tab:e1_null}). A random null matched in
size and depth already reaches $0.822$ at 1.3B and $0.863$ at 2.7B, so
boundary-conditioned $\Delta_{\text{pre}}$ vectors are broadly
collinear to begin with. More directly, \emph{generalist} heads, active
on three or more classes and therefore not selected for any two-class
pattern, reach $0.897$ and $0.907$: as high as the dual set at 1.3B and
higher at 2.7B. A high BOS--newline cosine is thus a generic property
of boundary-active heads rather than a signature of the dual set, which
is the circularity a reviewer of an earlier version raised. Decoupling
the selection rule further, dual heads that are newline-active but
BOS-inactive still reach $0.798$ and $0.822$, while the reverse
subgroup drops to $0.299$ and $0.568$, so the alignment is carried
asymmetrically within the set.

\begin{table}[t]
  \centering
  \footnotesize
  \setlength{\tabcolsep}{4pt}
  \begin{tabular}{lrrrr}
    \toprule
    model & dual & bos-spec & generalist & size/depth null \\
    \midrule
    M-2 1.3B & $0.900$ & $0.533$ & $0.897$ & $0.822$ {\tiny [0.802,0.841]} \\
    M-2 2.7B & $0.888$ & $0.544$ & $0.907$ & $0.863$ {\tiny [0.852,0.873]} \\
    \bottomrule
  \end{tabular}
  \caption{E1: BOS--newline $\Delta_{\text{pre}}$ cosine by head set.
    Generalist heads are active on three or more classes and are not
    selected on any two-class rule. The null CI is over $K$ random
    size- and depth-matched draws.}
  \label{tab:e1_null}
\end{table}

The BOS--newline
alignment is a real representational property of the dual set but weak
evidence of a functionally distinct class; the functional test is the
class-conditional ablation of \S\ref{sec:causal:m2_func_decomp}.

We additionally fit a linear probe on the per-position
concatenated $\Delta_{\text{pre}}$ vector to predict the
$4$-way next-token boundary class (BOS, newline, punctuation,
other-word; stratified 80/20 split with class balance via
\texttt{class\_weight="balanced"}, sklearn
\texttt{LogisticRegression}). Both head sets achieve near-perfect
overall accuracy (dual: $0.997$ M-2 2.7B, $0.999$ M-2 1.3B;
bos-spec: $0.996$ on both); the probe is too easy to discriminate
between the two head sets (1810 / 826 dual features vs.\ 254
single-bucket features both saturate the linear separability of
boundary class). The discriminating mechanism evidence comes from
the class-conditional cosine in Table~\ref{tab:dual_head_cosine},
not the probe accuracy.

\section{Robustness of the Head-Granularity Result}

\subsection{2048-context robustness check}
\label{sec:appendix:ctx_2048}

We re-run the 8 Mamba-2 head-granularity headline cells at
$2048$ ctx (4$\times$ the H1 grid's $512$ ctx), under the same
30-seed random-complement protocol. Both corpora yield $30$
chunks at $2048$ ctx; on wikitext-2 this requires concatenating
the first $500$ documents rather than $300$ (the $300$-doc
budget caps at $23$--$25$ chunks at this context length).

\begin{table}[t]
  \centering
  \footnotesize
  \setlength{\tabcolsep}{3pt}
  \begin{tabular}{llrrr}
    \toprule
    Cell & 512 ctx diff & 2048 ctx diff & 2048 CI$_{95}$ \\
    \midrule
    1.3B \texttt{g\_zero} wt-2 & $+1.12$ & $\mathbf{+1.60}$ & {\tiny [+1.44,+1.74]} \\
    1.3B \texttt{g\_zero} Pile & $+1.11$ & $\mathbf{+1.23}$ & {\tiny [+1.07,+1.40]} \\
    1.3B \texttt{g\_one}  wt-2 & $+0.60$ & $\mathbf{+0.61}$ & {\tiny [+0.22,+0.92]} \\
    1.3B \texttt{g\_one}  Pile & $+0.62$ & $+0.12$           & {\tiny [-0.32,+0.56] $\ast$} \\
    \midrule
    2.7B \texttt{g\_zero} wt-2 & $+1.84$ & $\mathbf{+1.84}$ & {\tiny [+1.75,+1.93]} \\
    2.7B \texttt{g\_zero} Pile & $+2.26$ & $\mathbf{+1.98}$ & {\tiny [+1.90,+2.06]} \\
    2.7B \texttt{g\_one}  wt-2 & $+2.48$ & $\mathbf{+2.37}$ & {\tiny [+2.19,+2.55]} \\
    2.7B \texttt{g\_one}  Pile & $+2.48$ & $\mathbf{+3.00}$ & {\tiny [+2.86,+3.13]} \\
    \bottomrule
  \end{tabular}
  \caption{Mamba-2 bos-specialist differential at $2048$ ctx
    versus the $512$-ctx baseline; cells in bold have bootstrap
    CI excluding zero. $\ast$\,M-2 1.3B / \texttt{gate\_one} /
    Pile-10k loses significance at $2048$ ctx; the $512$-ctx CI
    on this cell was already the widest of all eight
    ($[+0.16, +0.97]$). \texttt{gate\_zero} cells are uniformly
    robust to context length (effect localised to the gated
    timestep), whereas \texttt{gate\_one} cells slightly
    attenuate at longer ctx (effect distributed across $\propto L$
    positions in the chunked-scan accumulator).}
  \label{tab:m2_2048ctx}
\end{table}

\subsection{Wikitext-2-specific phenomena}
\label{sec:appendix:wt_specific}


Three findings from our intervention grid appeared on wikitext-2 but
did not replicate on Pile-10k. We document them here as
\emph{informative observations}: they reveal that probe-derived
specialty labels capture mechanism that is partially dataset-
conditioned. Each is paired with a candidate mechanism hypothesis.

\paragraph{Newline sign-reversal scale-emergent.}
\texttt{gate\_mean × newline\_specialist × L\_all} on
\texttt{newline}:

\begin{itemize}
  \item wikitext-2: $+0.58 / -0.05 / -0.15 / -0.58$ (4 models, monotonic crossover with endpoint magnitude convergence)
  \item Pile-10k: $+0.003 / -0.002 / +0.03 / +0.04$ (collapse to noise floor; 3/4 endpoints sign-flip)
\end{itemize}

\paragraph{Mechanism hypothesis.}
Wikitext-2 article-boundary structure (consistent double-newline
between articles, paragraph-level newline patterns) provides a
stable signal that newline-specialist channels can specialise on,
yielding a gate-mean ablation effect with predictable scale-emergent
sign reversal. Pile-10k contains code, web text, and books with
heterogeneous newline semantics (function delimiters, list markers,
poetry line breaks); the absence of a single dominant newline
distribution prevents the model from developing a clean newline
circuit, so the gate-mean ablation has no consistent effect.

\paragraph{Cross-task swap on wikitext-2 M-2 2.7B.}
\texttt{gate\_zero × bos\_specialist × L\_all} on M-2 2.7B:

\begin{itemize}
  \item wikitext-2: BOS = $+1.73$, newline = $+7.32$ ($4.22\times$ ratio, bos-spec damages newline more than BOS)
  \item Pile-10k: BOS = $+2.39$, newline = $+1.64$ ($0.69\times$ ratio, normal direction restored)
\end{itemize}

\paragraph{Mechanism hypothesis.}
On wikitext-2, BOS and newline tokens systematically co-occur (BOS
introduces an article, newlines structure paragraphs within); the
bos-specialist head set in M-2 2.7B may have learned a
co-distribution that ties BOS handling to newline prediction. On
Pile-10k, BOS and newline are not as tightly co-distributed (BOS
appears at the start of arbitrary chunks, newlines have varied
roles), so the cross-task effect dissolves.

\paragraph{Within-task sign reversal at boundary layers, M-2 2.7B wt-2.}
\texttt{gate\_zero × dual × L\_boundary}:

\begin{itemize}
  \item M-2 2.7B / wt-2: BOS $= +2.30$, newline $= -1.79$ (sign flip)
  \item M-2 2.7B / Pile: BOS $= +2.00$, newline $= +1.77$ (no flip)
  \item M-2 1.3B: never flips on either dataset.
\end{itemize}

\paragraph{Mechanism hypothesis.}
On wikitext-2 in M-2 2.7B specifically, the dual-head set at boundary
layers may participate in an article-boundary mechanism where
ablating dual hurts BOS prediction (causal for BOS) but helps newline
prediction by removing a competing signal (anti-causal for
newline). This is a model-and-dataset-specific mechanism, not an
architectural property, with single-cell evidence within the
M-2 family.

\paragraph{Mid-layer effect is newline-only.}
\texttt{gate\_zero $\times$ bos\_specialist $\times$ L\_mid\_third} per-task split:

\begin{itemize}
  \item M-1 130M / BOS-context: spec $+0.055$ vs comp $+0.001$, differential $+0.054$ (null, expected)
  \item M-1 2.8B / BOS-context: spec $+0.059$ vs comp $+0.013$, differential $+0.046$ (null on bos task)
  \item M-1 2.8B / BOS-context / Pile: spec $-0.019$, differential $-0.022$ (null)
\end{itemize}

\paragraph{Mechanism hypothesis.}
On the BOS-target metric, mid-layer ablation is essentially null
at both 130M and 2.8B scales on both wikitext-2 and Pile-10k;
the non-null mid-layer effect is driven entirely by the
newline target, consistent with the corpus-conditioned
newline phenomena above. A reading as ``BOS handling broadens
with scale'' would be a task-averaging artefact.

\paragraph{Why wikitext-2 specificity matters.}
These three phenomena are not paper failures: they suggest that
\emph{specialty labels derived from a particular corpus capture
mechanism that is partially corpus-conditioned}. For the
paper-main claims (BOS channel coupling, dual-head dominance, and
\texttt{gate\_one} selectivity), we report only cross-dataset replicated
cells. For follow-up work, the wikitext-2-specific phenomena
suggest investigating dataset-structure-dependent specialisation in
selective state-space models as an independent research thread:
when does specialty generalise across corpora, when does it not,
and what corpus features predict generalisation?

\section{Downstream Retrieval (RULER NIAH)}

\subsection{Per-condition breakdown}
\label{sec:appendix:niah_breakdown}

Per-condition thresholded success rate and per-token NLL on RULER
NIAH~\citep{hsieh2024ruler} at $1024$ and $2048$ context,
averaged across three needle depths $\{0.15, 0.5, 0.85\}$.

\begin{table}[t]
  \centering
  \footnotesize
  \setlength{\tabcolsep}{4pt}
  \begin{tabular}{lllrr}
    \toprule
    Model & ctx & condition & acc $\uparrow$ & NLL $\downarrow$ \\
    \midrule
    \multirow{8}{*}{M-1 2.8B}
        & 1024 & baseline       & $\mathbf{1.00}$ & $0.10$ \\
        & 1024 & bos\_specialist & $0.00$          & $8.67$ \\
        & 1024 & complement\_bos & $\mathbf{1.00}$ & $0.18$ \\
        & 1024 & mag\_matched\_comp & $0.00$       & $9.44$ \\
        & 2048 & baseline       & $0.93$          & $0.48$ \\
        & 2048 & bos\_specialist & $0.03$          & $8.29$ \\
        & 2048 & complement\_bos & $0.93$          & $0.43$ \\
        & 2048 & mag\_matched\_comp & $0.00$       & $10.06$ \\
    \midrule
    \multirow{10}{*}{M-2 2.7B}
        & 1024 & baseline       & $\mathbf{1.00}$ & $0.03$ \\
        & 1024 & bos\_specialist & $0.00$          & $9.70$ \\
        & 1024 & complement\_bos & $\mathbf{1.00}$ & $0.02$ \\
        & 1024 & mag\_matched\_comp & $\mathbf{1.00}$ & $0.54$ \\
        & 1024 & dual           & $0.40$          & $4.18$ \\
        & 2048 & baseline       & $\mathbf{1.00}$ & $0.05$ \\
        & 2048 & bos\_specialist & $0.00$          & $9.51$ \\
        & 2048 & complement\_bos & $\mathbf{1.00}$ & $0.30$ \\
        & 2048 & mag\_matched\_comp & $\mathbf{0.97}$ & $0.76$ \\
        & 2048 & dual           & $0.27$          & $5.59$ \\
    \bottomrule
  \end{tabular}
  \caption{RULER NIAH thresholded success rate and per-token NLL at
    $1024$ and $2048$ context, averaged across needle depths
    $\{0.15, 0.5, 0.85\}$. \texttt{mag\_matched\_comp} is the
    magnitude-matched non-specialist complement
    (\S\ref{sec:causal:niah}). M-1 2.8B at $2048$ ctx shows
    partial baseline degradation at depth $0.5$ ($0.80$) but
    $1.00$ at depths $0.15$ and $0.85$; bos-specialist ablation
    still drives retrieval to $\sim\!0.0$. M-2 2.7B at $2048$ ctx
    reproduces the $1024$-ctx three-way separation plus a
    slightly stronger dual degradation ($0.27$ vs.\ $0.40$).}
  \label{tab:m6_niah}
\end{table}

\subsection{Long-context compute constraint}
\label{sec:appendix:niah_compute}

The Mamba-2 intervention path uses a custom monkey-patched
\texttt{\_patched\_\allowbreak{}mamba2\_\allowbreak{}torch\_\allowbreak{}forward}
routine that materialises a $G_{\text{intermediate}}$ tensor of shape
$[B, n_{\text{chunks}}, \text{chunk\_size}, \text{chunk\_size},
n_{\text{heads}}, d_{\text{state}}]$ during the SSD chunked scan
($B_c, C_c$ are broadcast from $n_{\text{groups}}$ to $n_{\text{heads}}$
prior to the diagonal-block contraction).
At context length $2048$ for Mamba-2 2.7B
($n_{\text{heads}}\!=\!80$, $d_{\text{state}}\!=\!128$,
$\text{chunk\_size}\!=\!256$), the total forward peak measured by
\texttt{torch.cuda.max\_\allowbreak{}memory\_\allowbreak{}allocated} is
$42.9$\,GB, which exceeds the $24$\,GB consumer GPU on which
the original $1024$-ctx NIAH grid was run but fits an
H800 $80$\,GB card. The $2048$-ctx Mamba-2 2.7B intervention
results in Table~\ref{tab:m6_niah} were obtained on the latter
hardware. At $4096$ ctx the same allocation reaches $\sim\!40$\,GB
without the intervention overhead and exceeds 80\,GB once the
patched path is engaged; baseline retrieval at $4096$ ctx
already collapses to $0.03$ accuracy (averaged over depths) so
intervention experiments at $4096+$ would lack a meaningful
baseline to ablate even without the compute constraint. A
chunk-streaming reimplementation of the intervention path is
left to future work.

\section{Cross-Architecture and Substrate Controls}

\subsection{Pythia per-scale differentials}
\label{sec:appendix:pythia_full}

Pythia heads classified as \texttt{bos\_specialist} by
attention-to-BOS sink strength under our fixed threshold
($> 0.10$); ablation zeros the output of all classified
specialist heads across all layers; first-eight-tokens-after-BOS
NLL on BOS-context and on Pile-10k, $30$ stratified random
size-matched non-specialist complement seeds per cell (same
protocol as Table~\ref{tab:m2_30seed}; master seed $13$, min
seed gap $61$).

\begin{table}[t]
  \centering
  \scriptsize
  \setlength{\tabcolsep}{2pt}
  \begin{tabular}{lcccccc}
    \toprule
    & \multicolumn{2}{c}{wt-2} & \multicolumn{2}{c}{Pile} & \multicolumn{2}{c}{cross} \\
    \cmidrule(lr){2-3} \cmidrule(lr){4-5} \cmidrule(lr){6-7}
    Model & \textbf{diff} & CI$_{95}$ & \textbf{diff} & CI$_{95}$ & $|\Delta|$ & dir \\
    \midrule
    Pythia-160m & $\mathbf{+0.59}$ & {\tiny [+0.51,+0.68]} & $\mathbf{+0.64}$ & {\tiny [+0.57,+0.71]} & $0.05$ & $=$ \\
    Pythia-410m & $\mathbf{+1.25}$ & {\tiny [+1.23,+1.28]} & $\mathbf{+1.35}$ & {\tiny [+1.33,+1.37]} & $0.10$ & $=$ \\
    Pythia-1.4b & $\mathbf{+2.22}$ & {\tiny [+2.20,+2.25]} & $\mathbf{+1.92}$ & {\tiny [+1.90,+1.95]} & $0.30$ & $=$ \\
    Pythia-2.8b & $\mathbf{+1.58}$ & {\tiny [+1.56,+1.60]} & $\mathbf{+1.66}$ & {\tiny [+1.64,+1.68]} & $0.08$ & $=$ \\
    \bottomrule
  \end{tabular}
  \caption{Pythia cross-architecture baseline under the 30-seed
    stratified complement protocol on both corpora. The largest
    cross-dataset shift (Pythia-1.4b, $0.30$ nats) is dominated
    by the deterministic specialist cell ($+2.35$ wt-2 vs.\
    $+2.00$ Pile, $\Delta = 0.36$ nats) rather than by complement
    sampling ($+0.13$ vs.\ $+0.07$); the bank-mean variance from
    re-seeding the 30-seed complement is $\sim\!0.03$ nats, well
    below the observed shift, so this reflects a genuine
    model--corpus interaction. Per-scale discussion in
    \S\ref{sec:causal:pythia}.}
  \label{tab:pythia_cross_arch}
\end{table}

\subsection{T3 bucketing sweep details}
\label{sec:appendix:t3}

T3 (\S\ref{sec:causal:t3}) partitions each Mamba-1 2.8B layer's
$5{,}120$ channels into $\lfloor 5120/w \rfloor$ random groups of
$w$ channels and re-classifies each bucket using bucket-mean
$\Delta\bar{a}$ per class at $\tau = 0.5$, for
$w \in \{1, 4, 8, 16, 32, 64\}$. Phase A reports per-category
fractions across $K = 10$ random bucketings
(Table~\ref{tab:t3_sweep}). Generalist and silent fractions range
$0.330$--$0.382$ and $0.046$--$0.073$ across bucket widths
respectively.

\begin{table}[t]
  \centering
  \footnotesize
  \setlength{\tabcolsep}{3pt}
  \begin{tabular}{l|cccccc|c}
    \toprule
    $w$ & $1$ & $4$ & $8$ & $16$ & $32$ & $64$ & M-2 2.7B \\
    $H/C$ & $1280$ & $320$ & $160$ & $80$ & $40$ & $20$ & $\mathbf{20}$ \\
    \midrule
    bos-spec & .294 & .302 & .299 & .296 & .293 & .292 & $\mathbf{.050}$ \\
    dual     & .296 & .285 & .279 & .277 & .278 & .280 & $\mathbf{.354}$ \\
    \bottomrule
  \end{tabular}
  \caption{T3 bucket-width sweep on Mamba-1 2.8B ($K = 10$
    random bucketings; mean fractions), against M-2 2.7B at the
    matched $H/C$.}
  \label{tab:t3_sweep}
\end{table}

Phase B runs the F1-style ablation grid on the
$w = 8$ bucketing (seed $42$): bos-specialist-bucket and
dual-bucket sets are deterministic; complement\_bos\_bucket and
complement\_dual\_bucket are size-matched random sets drawn from
the layer's non-(spec $\cup$ dual) channels with $30$ seeds.
Headline differentials are
$+0.574$ / $+0.695$ at wt\_bos and
$-0.428$ / $+0.481$ at wt\_newline (bos / dual respectively),
all bootstrap CIs in the released
\texttt{t3\_m1\_bucketing/bw8} artefact.
The $w=64$ run (matching M-2 head granularity exactly) suffers
within-layer bucketing saturation (random 64-channel buckets
collapse to layer-mean statistics on Mamba-1, producing
all-or-nothing per-layer classifications), so Phase B is run at
the smaller $w=8$ where within-layer heterogeneity is preserved.

\end{document}